\documentclass[letter,11pt]{article}

\usepackage[ margin=1in]{geometry}
\usepackage{times}
\usepackage{graphicx}
\usepackage{wrapfig}
\usepackage{subcaption}
\usepackage{setspace}
\usepackage{multicol}
\usepackage{comment} 
\usepackage[T1]{fontenc}
\usepackage[normalem]{ulem}
\usepackage{enumitem}
\usepackage{amsmath}
\usepackage{amssymb}
\usepackage{fancyhdr}
\usepackage{lastpage}
\usepackage{hyperref}
\usepackage{apacite}
\usepackage{xcolor}

\title{Reframing linguistic bootstrapping as joint inference using visually-grounded grammar induction models}
\date{Eva Portelance$^{1,3}$\thanks{Corresponding author: eva.portelance@hec.ca}, Siva Reddy$^{2,3}$, Timothy J. O'Donnell$^{2,3}$ \\
\hfill \\
$^1$HEC Montréal \\
$^2$McGill University\\
$^3$Mila - Quebec AI Institute \\
\hfill \\
June 2024}
\begin{document}

\maketitle

\begin{abstract}
Semantic and syntactic bootstrapping posit that children use their prior knowledge of one linguistic domain, say syntactic relations, to help later acquire another, such as the meanings of new words. Empirical results supporting both theories may tempt us to believe that these are different learning strategies, where one may precede the other. Here, we argue that they are instead both contingent on a more general learning strategy for language acquisition: joint learning. Using a series of neural visually-grounded grammar induction models, we demonstrate that both syntactic and semantic bootstrapping effects are strongest when syntax and semantics are learnt simultaneously. Joint learning results in better grammar induction, realistic lexical category learning, and better interpretations of novel sentence and verb meanings. Joint learning makes language acquisition \textit{easier} for learners by mutually constraining the hypotheses spaces for both syntax and semantics. Studying the dynamics of joint inference over many input sources and modalities represents an important new direction for language modeling and learning research in both cognitive sciences and AI, as it may help us explain how language can be acquired in more constrained learning settings.
\end{abstract}

\section{Introduction}
With the advent of large language models (LLMs) it has become hard to deny that, if you have little to no limitations on the quantity of input data, amount of computation, or and memory size, much linguistic structure can be learnt \cite{mahowald2023dissociating,piantadosi2023modern}. Clearly human learners are not as unconstrained and therefore LLMs cannot be said to mimic human language learning. However, in light of the fact that these models can learn what they do without the need for language-specific learning mechanism or innate linguistic knowledge, it is important to reassess existing debates on language acquisition in children.

Language models have been proposed as useful tools for building proof-of-concept arguments for what is learnable from linguistic input (\citeNP[ch. 1.2]{lappin2021deep}, \citeNP{tsuji2021scala, warstadt2022what, pearl2023computational, portelance2023roles}). As such, they can inform innateness debates \cite{clark2011linguistic, crain2012syntax}, which try to establish how much innate knowledge is necessary to learn language and how specific to language learning strategies must be for its acquisition. Proofs of concept are however limited to informing us of what can possibly be learnt under the learning conditions of the target model. \citeauthor{portelance2023roles} (2023) suggest that a more interesting way to use these language models is for hypothesis generation, whereby models are used to study the dynamics at play during language learning to propose novel theories about the strategies which drive language acquisition in people. This paper adopts this approach, proposing a unified explanation for linguistic bootstrapping phenomena.

Linguistic bootstrapping theories posit that children use their prior knowledge in one linguistic domain, for example syntactic relations, to help with the acquisition of another, such as the meanings of new words. Here, we will address two theories, semantic bootstrapping and syntactic bootstrapping. These proposals come from a theoretical landscape which assumed that learning is algorithmic or stage-based in nature, requiring some prior innate linguistic knowledge as a starting point. The bootstrapping debates have then centered on the question \textit{what linguistic knowledge do we start from?} and \textit{how do children use this knowledge to bootstrap new knowledge and eventually acquire language?} \cite{grimshaw1981form,pinker1984language, landau1985language, gleitman1990structural}.

\subsection{Linguistic bootstrapping debates}

Though linguistic bootstrapping can extend to all levels of linguistic representation, from phonology to pragmatics, we will concentrate on the two types which have been most discussed, semantic bootstrapping and syntactic bootstrapping.

Broadly, in semantic bootstrapping proposals, children are said to use their knowledge of semantics and meaning to \textit{bootstrap} syntactic knowledge. The most famous formulation comes from \citeNP{pinker1984language, pinker2009language}. In this version, semantic bootstrapping is envisioned as an early language learning strategy which allows children to learn syntactic primitives such as syntactic categories (noun, verb, adjective...), by mapping them to early perceptual or cognitive categories\footnote{Somewhat confusingly, it is called \textit{semantic} bootstrapping, even though in its original formulation, it never refers to any knowledge of formal semantics or linguistic meaning, but instead to perceptual concepts independent of language.} (individual, action, state...). For example, children may induce a syntactic category like noun by noticing that there are words which name the semantic perceptual categories of persons or things. Once categories are learnt, a more symbiotic exchange would then emerge between syntactic and semantic knowledge to acquire new structures and word meanings. There is indirect empirical evidence which supports this hypothesis, such as the observation that children's first nouns correspond to physical objects, first verbs to actions, and first adjectives to perceptually salient attributes. Pinker describes semantic bootstrapping as a form of distributional learning, stating that ``children always give priority to distributionally based analyses, and [semantic bootstrapping] is intended to explain how the child knows which distributional contexts are the relevant ones to examine.''\cite[p.42-43]{pinker2009language} In this view, language learning can be thought of as a probabilistic mapping problem. Semantic bootstrapping is then a theory that tries to explain how children use aspects of meaning to acquire core pieces of syntactic structure. 

\begin{figure}
    \centering
    \includegraphics[height=4.7cm]{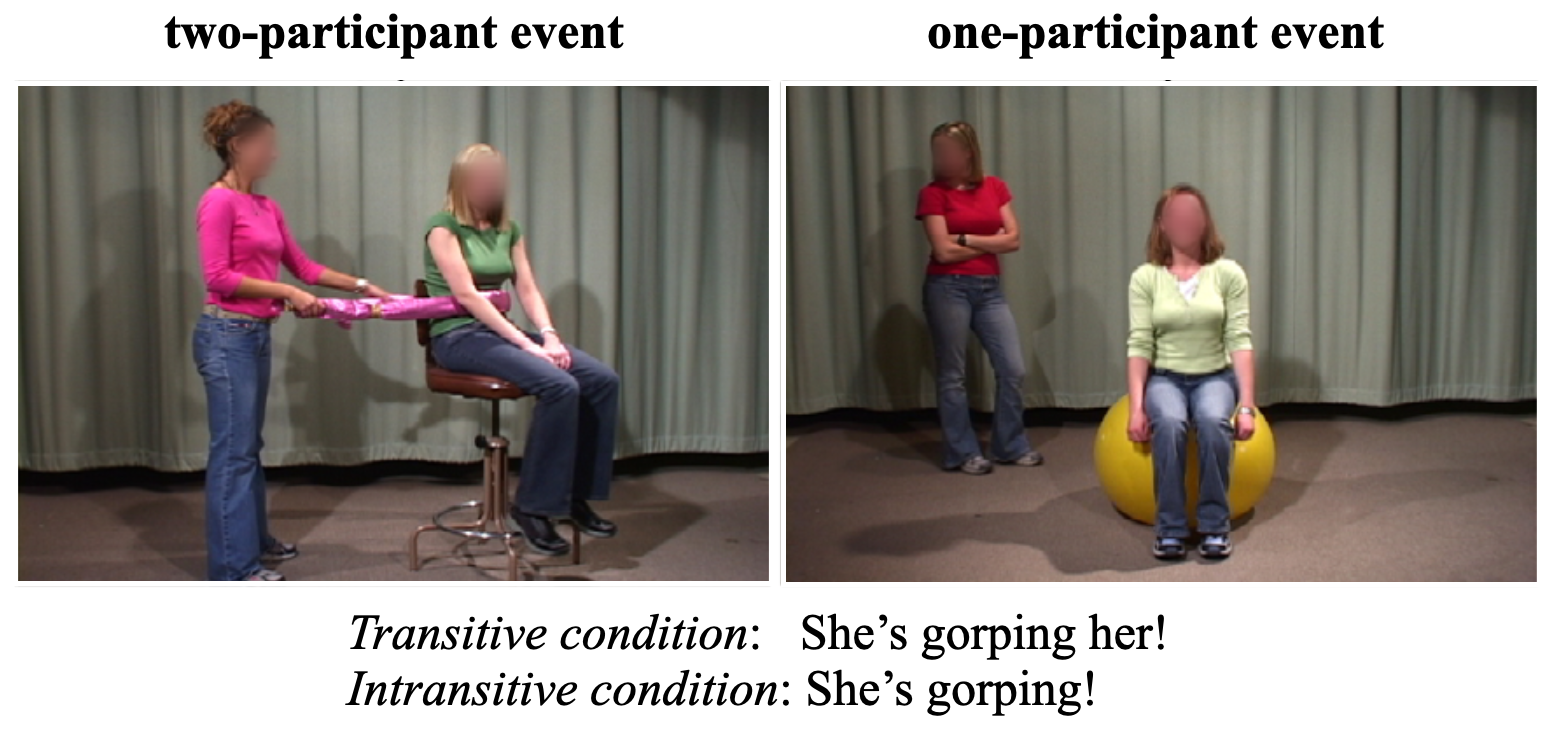}
    \caption{Example of nonce verb learning experimental paradigm from \protect\cite{yuan2012counting}, demonstrating indirect evidence for the syntactic bootstrapping hypothesis.}
    \label{fig:ex-exp}
\end{figure}

Syntactic bootstrapping proposals instead describe processes which involve the use of syntactic knowledge to \textit{bootstrap} new meanings and semantic knowledge. The proposal and first full description of this theory is associated with \citeNP{gleitman1990structural}. Gleitman makes the case that the same problems which plague syntax acquisition and have motivated many syntactic theories, also exist for vocabulary acquisition: the hypothesis space over syntactic structure and word meanings is simply too vast and must be limited in some way by prior knowledge or learning strategies to account for children's language learning abilities. As a way to limit this hypothesis space for vocabulary acquisition, she posits that learners must start with ``sophisticated presuppositions about the structure of language''. In other words, children start with some syntactic knowledge to bias their acquisition of new word meanings. Discussion of syntactic bootstrapping have for the most part been limited to the acquisition of verb meanings, though in principle it may extend to other types of words \cite{fisher2010syntactic}. There is once again indirect empirical evidence which supports this hypothesis, such as children's ability to infer a novel verbs meaning based on its argument structure \cite{naigles1990children}. For example, if a child is exposed to the transitive condition in Figure \ref{fig:ex-exp} and they have learnt to identify such argument structures, they can infer the presence of an agent (someone performing an action) and a patient (someone upon whom an action is performed) are likely necessary in the correct visual interpretation, picking the two-participant event---and more generally identifying and learning the meaning of `gorping'. In its original strong form, syntactic bootstrapping requires innate language-specific abstract knowledge of syntactic structures, which is independent of word meaning or grounding. More recent descriptions however present early abstract structural knowledge as possibly probabilistic and learnt from more primitive concepts \cite{fisher2020developmental}.

Though sometimes erroneously characterised as opposing theories, semantic and syntactic bootstrapping are complementary theories. Their characterization as conflicting may stem from the conjecture that semantic bootstrapping leads to a lexically-driven view of language acquisition where language is learnt by mapping words to meanings and syntactic primitives (bottom-up perspective), while syntactic bootstrapping leads to a more abstraction-based view where language is learnt by mapping sentential structures to word meanings (top-down perspective). However, these theories can very much coexist. In \citeNP{pinker2009language}'s semantic bootstrapping proposal, the second stage of learning after primitive syntactic categories are learnt is called ``structure-dependent distributional learning" and is akin to syntactic bootstrapping. In their review article on syntactic bootstrapping, \citeauthor{fisher2020developmental} (2020) articulate the leading view: that there are independent learning strategies which both rely on the same tight bond between syntax and semantic representation.
In either case, one theory does not exclude the other; both semantic and syntactic bootstrapping can coexist.

\subsection{Grammar induction models and linguistic bootstrapping}

Grammar induction is the task of learning a grammar -- or set of rules and structures -- given a corpus of sentences. In the past, it had proven quite difficult to do using purely statistical learning or distributed models, especially for natural language which requires more expressive grammars, such as probabilistic context-free grammars (PCFG; \citeNP{klein2004corpus,klein2005natural,cohn2010inducing, perfors2011learnability}). At the time, the solution had been to use probabilistic models with more informed prior knowledge of language, such as access to categories, head-branching bias or semantic mappings \cite{chater2006probabilistic, muralidaran2020systematic}. One such model of particular interest to this paper was the semantic bootstrapping model from \citeNP{abend2017bootstrapping}, which was an instantiation of the theory by the same name. The authors proposed a Bayesian grammar induction model which learnt both syntactic derivations for sentences and word meanings conditioned on knowing sentential meanings (given in the form of compositional semantic derivations). Using their model, the authors argued that syntactic bootstrapping is an emergent effect which follows from this tight bond imposed on syntax and semantics during semantic bootstrapping. In such a proposal, syntactic bootstrapping is not a learning strategy in its own right, but an effect which follows from semantic bootstrapping strategies.

More recently, access to better computational resources and neural network architectures have lead to significant advances for grammar induction models -- as with language models. Researchers have managed to design successful distributional approaches to natural language grammar induction without the need for overly informative priors \cite{kim-etal-2019-compound, drozdov-etal-2019-unsupervised-latent}. These models have since been augmented by introducing visual-grounding, finding that access to visual information helps models induce grammars producing more accurate constituency trees \cite{shi-etal-2019-visually, zhao-titov-2020-visually, jin-schuler-2020-grounded,wan2022unsupervised} --- a result reminiscent of semantic bootstrapping. Taking inspiration from these recent advances, we present our own model which builds on this previous work and additionally learns to interpret visual information via access to induced trees. Thus, in addition to semantic bootstrapping, the model has the possibility of performing syntactic bootstrapping. Like \citeNP{abend2017bootstrapping}, it addresses both semantic and syntactic acquisition; being a neural network, we are able to introduce the added complication of learning not just grammar, but semantic representations from scratch as well. It thus more closely models young children's reality and allows us to better examine the dynamics behind joint learning of syntax and semantics.

\subsection{Our proposal}

Both semantic and syntactic bootstrapping theories are predicated on children learning \textit{something first} to bootstrap the acquisition of \textit{something later}. The empirical results which exist in both direction may tempt us to believe that these are different learning strategies, where one may possibly precede the other (i.e. children learn the meanings of simple nouns first or maybe they learn basic syntactic relations first). However, another possibility is that they are both contingent on a more general learning strategy for language acquisition.

In this paper, we make the following theoretical proposal: linguistic bootstrapping follows from joint learning over multiple levels of linguistic representation, via simultaneous access to multiple input modalities. We argue that neither syntactic nor semantic bootstrapping are independent learning strategies, as they have previously been presented, but both learning effects which arise as a consequence of a probabilistic joint learning strategy over both syntactic and semantic levels of representation for language. Furthermore, we propose that no prior linguistic knowledge is necessary beyond a bias towards learning abstract categories, to acquire both syntactic and semantic representations.

Learning syntactic structures can facilitate learning to meanings, and conversely, learning meanings can facilitate learning syntactic structures. In a joint inference process over both syntactic and semantic representations, each hypothesis space can constrain the other and help learners to simultaneously acquire syntax and semantics.

In the sections which follow, we will use a neural visual grammar induction model to show how constraints during joint inference on one linguistic domain can affect another and lead to better generalisation in completely novel contexts\footnote{All data, code, analyses and experimental results are publicly available at \hyperlink{github.com/evaportelance/structure-meaning-learning}{github.com/evaportelance/structure-meaning-learning} .
}. Our model learns grounded representations of both syntactic structure and semantic meanings from sentence-image pairs using a statistical learning algorithm. We ask the following questions:  \textbf{1.} (analogous to semantic bootstrapping) Can access to visual-grounding and the ability to learn semantic representations in a joint learning setting facilitate learning grammars that generalize better to unseen contexts? \textbf{2.} (analogous to syntactic bootstrapping) Can access to linguistic structure and the ability to learn grammar in a joint learning setting facilitate learning and interpreting novel words and contexts?


\section{The dataset}\label{sec:data}
\begin{figure}
    \centering
    \includegraphics[width=\textwidth]{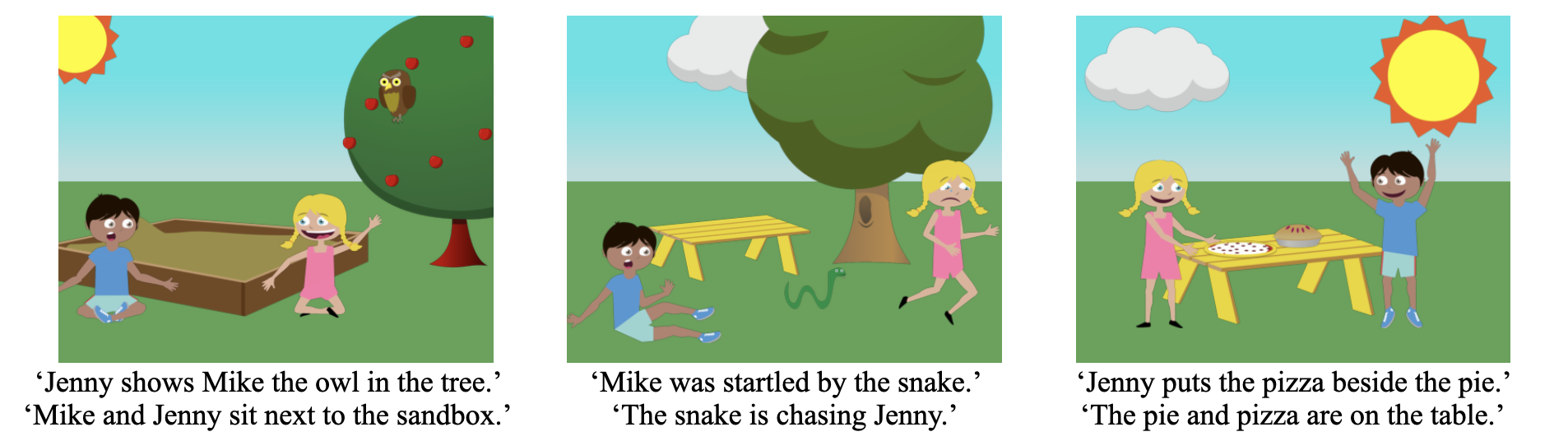}
    \caption{Examples image-sentence pairs from the Abstract Scenes dataset \protect\cite{zitnick2013bringing, zitnick2013learning}}
    \label{fig:data-ex}
\end{figure}

For all of the demonstrations and experiments reported in this paper, we use the \href{http://optimus.cc.gatech.edu/clipart/}{Abstract Scenes dataset} \cite{zitnick2013bringing, zitnick2013learning}, a dataset composed of clip-art scenes meant to resemble children's book illustrations paired with simple sentences narrating the contents of the images, Figure \ref{fig:data-ex} contains examples of these image-sentence pairs.

Previous work on visually-grounded grammar induction has used image-caption pairs from datasets like MS-COCO \cite{lin2014microsoft, chen2015microsoft}; the problem with image captions is that they are not always complete sentences, but often complex noun phrase descriptions which lack main verbs, for example ``A man sitting on a park bench with an umbrella''. For this work, it was important to use image-sentence pairs with complete sentences containing main verbs. Much of the existing literature on linguistic bootstrapping, especially syntactic bootstrapping, has centered on the acquisition of verb meaning using argument structure. Furthermore, not having access to main verbs enough may bias grammar induction models towards trees that favor different words as sentence heads.

The images and sentences from Abstract Scenes contain depictions and descriptions of actions with both transitive and intransitive main verbs that lend themselves well to syntactic and semantic bootstrapping experimentation. This dataset has also been used before in modeling experiments for language acquisition research \cite{nikolaus-fourtassi-2021-evaluating, nikolaus-fourtassi-2021-modeling}. In total it contains 10,020 images each paired with 6 different sentences, for a total of 60,160 image-sentence pairs.

\subsection{Test-train splits}
The Abstract Scenes Dataset does not come with preexisting data splits, so we designed our own to evaluate syntactic and semantic bootstrapping on in-distribution syntactic structure learning and out-of-distribution verb learning.

For the out-of-distribution experiment presented in section \ref{sec:syn-boot}, we created a test split that contained image-sentence pairs with novel main verbs. In other words, if the model was trained on instances of \textit{throw, kiss, cry} it was then evaluated on \textit{toss, hug, smile}. The latter verbs can be considered nonce verbs from the perspective of the model. To create our test data, we first extracted all of the main verbs with more than 5 instances in the dataset (around 480 verbs), grouped them by verb stem (280 unique verb stems), annotated them for transitivity and object animacy. We then hand selected 10 intransitive verb stems and 10 transitive stems taking animate objects and 10 taking inanimate objects. We selected these verbs such that they appeared in varied sentential contexts (i.e. there were no synomyms or closely related verbs) and such that the total number of transitive and intransitive sentences were as close as possible. The training and test verb stem lists are available in Appendix \ref{app:data}. In total, the test set contained 1708 sentences (718 transitive, 990 intransitive) with 30 different held out verb stems (10 transitive-animate, 10 transitive-inanimate, 10 intransitive).

For experiments in sections \ref{sec:sem-boot} with in-distribution tests, we simply reintroduced half of the test set to training, so that all verbs were now seen at least once during training, resulting in 833 test sentences (357 transitive, 476 intransitive) with 30 different verb stems. Thus, test sentences in these in-distribution evaluation were novel sentences containing known verbs.


\section{The joint-learning model}

Our joint-learning model is a visually-grounded grammar induction
model based on VC-PCFG (visually-grounded compound probabilistic context-free grammar) by \citeNP{zhao-titov-2020-visually}. It is a model that
combines two learning objectives: (1) inducing a compound
probabilistic context free grammar, or C-PCFG
\cite{kim-etal-2019-compound}, over a set of sentences, which we will
call the syntactic objective, and (2) maximizing the similarity
between predicted syntactic structure of those sentences with the
representations of the images (described below), which we will call the
semantic objective.

\subsection{The syntactic objective}

First, we define the type of grammar our model induces and the syntactic objective. C-PCFGs are extensions of probabilistic context free grammars (PCFGs) \cite{kim-etal-2019-compound}. A PCFG, $\langle \mathcal{G}, \mathbf{\pi}\rangle$  is a grammar $\mathcal{G}$ coupled with $\mathbf{\pi}$, a probability function over $\mathcal{G}$. Here, the context free grammar can be formalized as a 5-tuple of finite sets $\mathcal{G} = (S,\mathcal{N}, \mathcal{P}, \Sigma, \mathcal{R})$, where $S$ is the start symbol, $\mathcal{N}$ the nonterminal categories, $\mathcal{P}$ the preterminal categories, $\Sigma$ the vocabulary or set of terminals, and $\mathcal{R}$ the production rules over words and categories. The rules in $\mathcal{R}$ are in Chomsky normal form:
\begin{align*}
S\rightarrow A,&  &A\in \mathcal{N}, \\
A\rightarrow BC,&  &A\in \mathcal{N}, B, C\in\mathcal{N}\cup\mathcal{P},  \\
T\rightarrow w&, &T\in \mathcal{P}, w\in \Sigma\,.
\end{align*}

The probability function $\mathbf{\pi}$ assigns some non-negative value to every production rule $r \in \mathcal{R}$. In most PCFGs, it is defined as a set of categorical probability distributions 
where there is a separate categorical for every set of rules with the same left-hand side such that $\sum_{\alpha \mid A \rightarrow \alpha}\pi(A \rightarrow \alpha)=1$. However, as shown in \citeNP{kim-etal-2019-compound}, the strong context-free assumption instantiated by this type of probability function isn't conducive to effective grammar induction. Instead, we would like to have a way to share information across rule types. Their proposed solution is the C-PCFG, which instead assumes that rule probabilities $\mathbf{\pi}$ follow a compound distribution \cite{robbins1951}.\footnote{C-PCFGs rule probabilities are thus not strictly independent of one another but are all conditioned on $\mathbf{z}$. C-PCFGs are in fact a mixture of PCFGs, satisfying the context-free assumption when conditioned on some value for the random variable $\mathbf{z}$.}. Each rule probability is parameterized by embeddings $\bf{u}, \bf{w}$ and by a sentence-global latent parameter $\mathbf{z}$, where $\mathbf{z}$ is itself a random variable sampled from a spherical Gaussian with prior parameters $\gamma$. Intuitively, $\mathbf{z}$ represents shared information in each sentence that is accessible throughout its tree derivations at each rule application. The introduction of this latent variable $\mathbf{z}$ turns out to be crucial for successful grammar induction as it allows information to flow between parts of the tree, making what used to be a very hard problem solvable \cite{kim-etal-2019-compound}. Here, our C-PCFG, $\langle \mathcal{G}, \mathbf{\pi}\rangle$, defines a compound probability distribution over our grammar using the following process:

\begin{equation}
\mathbf{z} \sim \mathrm{SphericalGaussian}(\gamma), \\
\end{equation}

Thus, conditioned on both $\mathbf{z}$ and the rule types, rule probabilities are then given by the following function:

\begin{equation}
\pi_{\mathbf{z}}(r) = \left\{\begin{array}{lr}
        \frac{\exp(\mathbf{u}_{A}^\top f_s([\mathbf{w}_{S}; \mathbf{z}]))}
{\sum_{A'\in\mathcal{N}}\exp(\mathbf{u}_{A'}^\top f_s([\mathbf{w}_{S}; \mathbf{z}]))}, & \text{for } r \in S\rightarrow A\\
        \frac{\exp(\mathbf{u}_{BC}^\top [\mathbf{w}_{A}; \mathbf{z}])}
{\sum_{B',C'\in\mathcal{N}\cup\mathcal{P}}\exp(\mathbf{u}_{B'C'}^\top [\mathbf{w}_{A}; \mathbf{z}])}, & \text{for } r \in A\rightarrow BC\\
        \frac{\exp(\mathbf{u}_{w}^\top f_t([\mathbf{w}_{T}; \mathbf{z}]))}
{\sum_{w'\in\Sigma}\exp(\mathbf{u}_{w'}^\top f_t([\mathbf{w}_{T}; \mathbf{z}]))}, & \text{for } r \in T \rightarrow w
        \end{array}\right\}
\end{equation}

where $\mathbf{u}_{\alpha}, \mathbf{w}_{\alpha}$ are vector embeddings over all possible left and right hand sides of rules, $\alpha \in \{\{S\}, \mathcal{N}, \mathcal{P}, \Sigma, (\mathcal{N} \cup \mathcal{P}) \times (\mathcal{N} \cup \mathcal{P})\}$,  and $[\cdot ; \cdot ]$ is vector concatenation. $f_t(\cdot)$ and $f_s(\cdot)$ are multi-layer perceptron (MLPs) encoders which are used to encode root and terminal rules\footnote{Following \citeNP{kim-etal-2019-compound}, no such MLP is used with non-terminal rules. See Appendix \ref{app:model} for detailed description of these encoder architectures.}. For further motivation for the use of this compound distribution, we refer readers to the original paper \citeNP{kim-etal-2019-compound}.

To induce a grammar given a corpus $\mathcal{C}$, we want to maximize the log-likelihood of each sentence $s \in \mathcal{C}$. The likelihood of a sentence under our C-PCFG is the the sum of the probability of every possible tree derivation $t$ for $s$, conditioned on $\mathcal{G}$, some $\mathbf{z}$ and $\mathbf{\pi}_\mathbf{z}$.
\begin{equation}
    p_{\theta}(s | \mathbf{z}) = \sum_{t \in \mathcal{T}_{\mathcal{G}}(s)} \prod_{r \in t_{\mathcal{R}}} \pi_{\mathbf{z}}(r)
  \end{equation}

where $\theta$ represents the hyperparameters of our grammar model, $\mathcal{T}_{\mathcal{G}}(s)$ is the set of all derivations for $s$, and $t_{\mathcal{R}}$ the set of rules in a given tree $t$.

In order to get the log-likelihood of each sentence $s$ irrespective of $\mathbf{z}$, we need to take the following integral:
\begin{equation*}
    \log p_{\theta}(s) = \log \int_{\mathbf{z}} \sum_{t \in \mathcal{T}_{\mathcal{G}}(s)} \prod_{r \in t_{\mathcal{R}}} \pi_{\mathbf{z}}(r)p_{\gamma}(\mathbf{z}) d\mathbf{z}
\end{equation*}
However, computing this integral is intractable. Instead, we can estimate a local maximum log-likelihood of a sentence using variational inference and maximizing the evidence lower bound (ELBO):
\begin{equation}
\log p_{\theta}(s) \geq \text{ELBO}(s; \phi, \theta, \gamma) =
\mathbb{E}_{q_{\phi}(\mathbf{z} | s)}[\log p_{\theta}(s | \mathbf{z})] -
\text{KL}[q_{\phi}(\mathbf{z} | s) || p_{\gamma}(\mathbf{z})],
\end{equation}
where $q_{\phi}(\mathbf{z} | s)$ is our variational approximation of the posterior parameterized by $\phi$, in practice computed using another neural network\footnote{For the specifics of its implementation see Appendix \ref{app:model}}.

Thus, our syntactic objective is defined as:
\begin{equation}
\mathcal{L}_{\text{syntax}}(\mathcal{C}; \phi, \theta, \gamma) = - \sum_{s \in\mathcal{C}} \text{ELBO}(s; \phi, \theta, \gamma)
\end{equation}
It induces a C-PCFG,
$\langle \mathcal{G}, \mathbf{\pi}\rangle$, by estimating
the maximum log-likelihood of our corpus $\mathcal{C}$ using
variational inference. This approach amounts to turning the grammar induction problem into a parameter estimation problem, which we can do via gradient decent using our neural network implementation. In practice, we allow our grammar $\mathcal{G}$
to contain up to 30 non-terminal categories, 60 pre-terminals, and a
vocabulary of 2000 terminals/words\footnote{The number of categories
  is based on previous work
  \cite{kim-etal-2019-compound,zhao-titov-2020-visually}. We also
  tested other values (20,40) and did not find that it affected the
  results significantly. The vocabulary size was based on the number
  of unique words in our dataset, discounting typos. Other hyperparameters are available as the default settings in our code repository \hyperlink{github.com/evaportelance/structure-meaning-learning}{github.com/evaportelance/structure-meaning-learning}
  .}.

\subsection{The semantic objective}

Next, we define the procedure for determining the similarity between
the predicted syntactic structure of a sentence and its image and the
semantic objective. The semantic objective is based on the multimodal
contrastive learning approach used by many multimodal encoders,
including CLIP \cite{kiros2014unifying,radford2021learning}. These
approaches use a text encoder and visual encoder to generate
representations for captions and images that can be compared in a
shared visual-semantic space. However, like previous work on visually
grounded grammar induction models, instead of using a text encoder for
whole captions, we encode each constituent in a sentence independently and compare them to the image encoding --- or semantic representation
\cite{shi-etal-2019-visually,zhao-titov-2020-visually}. In practice, to ensure that the model is fully differentiable, we do this comparison for every possible constituent or word span and then weight their relative importance by their marginal likelihood under the induced grammar.

To extract a semantic representation, we first embed an image using
  the final layer of a pretrained convolutional neural network.
  However, unlike previous work that uses pretrained ResNet
  models---which are trained using supervised class labels for
  images---\cite{shi-etal-2019-visually,zhao-titov-2020-visually}, we
  train our own vision network on our dataset using an unsupervised
  learning algorithm. Given that our joint model serves as a cognitive
  model, we did this to prevent introducing any linguistic biases into
  the image representations which could result from exposure to object
  class labels during training. Instead, we trained a ResNet-50
  architecture from scratch on the Abstract Scenes images using a
  visual contrastive learning algorithm called SimCLR
  \cite{chen2020simple}. From this model we can extract
  the set of all image representations $\mathcal{V}$:

  \begin{equation*} \mathcal{V} = \texttt{net}_{\mathrm{image}}(\mathcal{I}),
  \end{equation*}
  where $\texttt{net}_{\mathrm{image}}$ is our pretrained
  self-supervised resnet-50 model and $\mathcal{I}$ our images.
  Finally, we then extract a semantic representation $\mathbf{m}$ for
  an image embedding $\mathbf{v} \in \mathcal{V}$ using our semantic
  encoder $f_{m}$, a trainable MLP:

  \begin{equation*} \mathbf{m} =
    f_{m}(\mathbf{v}). \end{equation*}
We make the distinction between image representations and semantic representations for the following reason: the image representations are pretrained and fixed during language learning, while the semantic representations are trainable and can be influenced by language. It may be though of as the distinction between visual perception, which is independent of language, and abstract representation and categorization of visual features, which may be influenced by language.

  As mentioned, for each image-sentence pair, our second learning
  objective tries to maximize the similarity between a semantic
  representation $\mathbf{m}$ for image content and the syntactic
  structures induced for sentences. In practice, we consider the
  similarity between a semantic representation of an image and all the
  constituent representations for all possible derivation trees of a sentence, which we afterwards way by their marginal likelihood. Thus, we define a matching objective
  for a single sentence-image pair $(s, \mathbf{m})$ as follows:
  \begin{equation}
    \ell_{\mathrm{pair}}(s,\mathbf{m}; \theta) = \sum_{c \in \mathrm{spans}(s)} \mathrm{match}(c, \mathbf{m}; \theta),
\end{equation}
where $\mathbf{c}$ is a possible constituent and $\mathrm{spans}(\cdot)$ is a function which returns all possible span of consecutive words from sentence $s$ of length $l$ where $1 < l < |s|$. As for our $\mathrm{match}(\cdot, \cdot)$ function, it is a weighted version of the contrastive learning hinge loss:
\begin{equation}
    \mathrm{match}(c, \mathbf{m}; \theta) = p_{\theta}(c|s, \mathbf{z})h(\texttt{biLSTM}(c), \mathbf{m}),
\end{equation}
where $p_{\theta}(c|s, \mathbf{z})$ is the marginal probability of a constituent across all possible $t \in \mathcal{T}(s)$\footnote{To see how we can derive this value from our derivation tree distribution $p_{\theta}(t | s, \textbf{z})$ over trees for $s$, see \citeNP{zhao-titov-2020-visually}.}, which weighs the hinge loss, $h(\texttt{biLSTM}(c), \mathbf{m})$, which compares each constituent independently encoded by a single-layered \texttt{biLSTM} language model, and the corresponding semantic representation for the sentence $\mathbf{m}$. This hinge loss is defined as follows, with $\mathbf{c}$ as the constituent encoding, following previous work \cite{shi-etal-2019-visually, zhao-titov-2020-visually}:
\begin{equation}\label{eq:hinge_loss}
h(\mathbf{c},  \mathbf{m}) =
\left[
\mathrm{sim}_{\mathrm{cos}}(\mathbf{c}', \mathbf{m}) - \mathrm{sim}_{\mathrm{cos}}(\mathbf{c}, \mathbf{m}) + \epsilon
\right]_{+}
+
\left[
\mathrm{sim}_{\mathrm{cos}}(\mathbf{c}, \mathbf{m}')\! -\! \mathrm{sim}_{\mathrm{cos}}(\mathbf{c}, \mathbf{m})\! +\! \epsilon
\right]_{+}
\,,
\end{equation}
where $[\cdot]_{+} = \max(0, \cdot)$, $\epsilon$ is a constant margin, and $\mathbf{c}', \mathbf{m}'$ are negative examples, or constituent and meaning representations from a different sentence-image pair. Intuitively, we want our model to learn that the similarity between matching constituent-image pair representations, $(\mathbf{c},  \mathbf{m})$, should be higher than for an unmatched one, for example $(\mathbf{c}',  \mathbf{m})$.

The semantic objective is then simply the sum of the pairwise objective across all sentence-image pairs.
\begin{equation}
    \mathcal{L}_{\text{semantics}} (\mathcal{C}, \mathcal{V}; \theta) = \sum_j \ell_{\mathrm{pair}}(s^{(j)},f_{m}(\mathbf{v}^{(j)}); \theta)
\end{equation}

\begin{figure}
    \centering
    \includegraphics[scale=0.5]{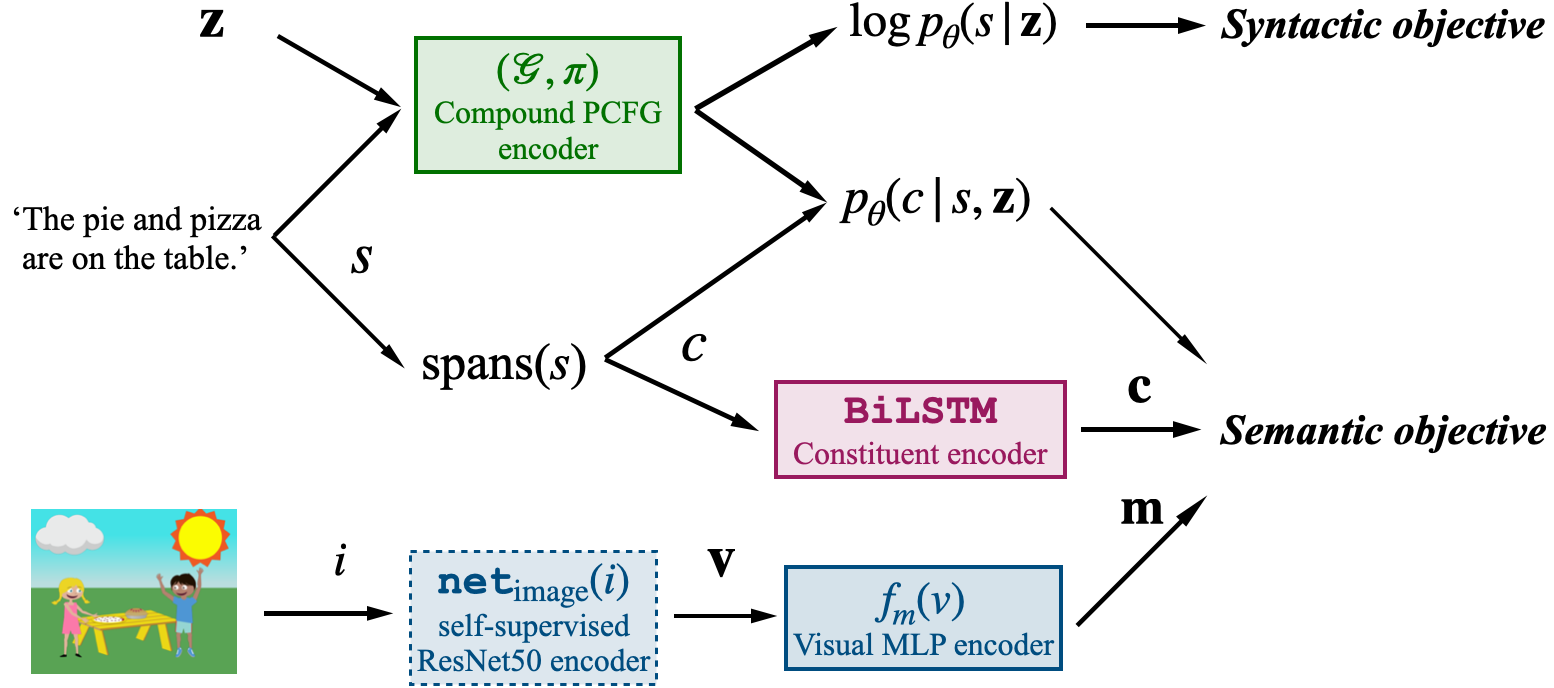}
    \caption{The joint model architecture}
    \label{fig:model}
\end{figure}

\paragraph{Joint-learning model: } during training, we optimize for both the syntactic objective and the semantic objective at the same time, using the joint loss :

\begin{equation}
    \mathcal{L}_{\text{joint}} (\mathcal{C}, \mathcal{V}; \phi, \theta, \gamma) = \alpha_1 \cdot \mathcal{L}_{\text{syntax}} (\mathcal{C}; \phi, \theta, \gamma) + \alpha_2 \cdot \mathcal{L}_{\text{semantics}} (\mathcal{C}, \mathcal{V}; \theta),
\end{equation}
where $\alpha_1, \alpha_2$ are constants, here both equal to 1. Since the syntactic and semantic objectives are interdependent, during learning they will affect the joint model's learning trajectory by mutually constraining updates to the grammar model and semantic encoder. On the one hand, the semantic objective will push the grammar model to favor rules which derive trees containing constituents that can be visually represented. On the other, the syntactic objective will determine the distribution over constituents being mapped to semantic space. We illustrate the model and this relation between objectives in Figure \ref{fig:model}.

In using our model as a cognitive model, we are making the following
assumptions about the acquisition of syntactic and semantic knowledge:
(1) we start with the prior knowledge that there are such things as
syntactic categories (though what they represent and how many there
are is unknown); (2) grammatical structure can be represented by a
PCFG; (3) meaning is grounded in visual representations. The first
assumption is feasable under most linguistic theories and theories of
acquisition. The second and third, are likely simplifications of
children's learning environment: PCFGs may not having enough
explanatory power for natural language
\cite{shieber1985evidence, huybregts1984weak}, and children ground language in embodied experiences going way
beyond still images. By definition, a model is a simplified
exemplification of a process. We acknowledge
that our model does not capture all the complexities present in a
child's naturalistic learning environment. However, we are using our model to study specific research questions and not language learning as a whole. We believe that this
joint-learning model can demonstrate how joint learning objectives for
language can aid language acquisition overall and explain
bootstrapping phenomena, like syntactic and semantic
bootstrapping.

\subsection{Model ablations and baselines}

To better understand the role that joint learning plays, we include
two ablated versions of the model in the demonstrations which follow:
the semantics-first model and the syntax-first model. In each case,
the model initially optimizes for a single learning objective and then
adds the second objective later during learning.

\paragraph{Semantics-first model: } This model starts by optimizing
for the semantic objective only. Since this model does not initially
have access to syntax or predicted structures, it uses the similarity
between the whole caption and images, as opposed to the similarity
between the constituents of a caption and an image\footnote{We also tried a version that uses the semantic loss defined above but where all constituents are equally likely and found that it performed exactly the same at test time as this simpler version.}. In other words, it
uses the traditional image-caption matching loss used in models like
CLIP \cite{radford2021learning}, having access to only the order in which words appeared but no structure beyond that. Half way through training, we add the syntactic objective, resulting in the regular joint-learning
objective. This model represents the condition where no syntactic
knowledge is initially acquired, or a ``semantics learnt first"
perspective.

\paragraph{Syntax-first model: } This model is initially trained
solely on the syntactic objective, or without visual grounding, in
other words using the original C-PCFG grammar induction objective from
\cite{kim-etal-2019-compound}. Half way through training, we add the
semantic objective, resulting once again in the regular joint-learning
objective. This second baseline represents the ``syntax learnt first"
perspective, where semantic knowledge is initially acquired before
learning syntactic structures.

Finally, we compare the joint-model to a target semantic model with
oracle knowledge of image content, which we call the visual-labels model.

\paragraph{Visual-labels model: } This model is trained using the same
objective as the joint-learning model, however, instead of using image
representations from our unsupervised pretrained image encoder, it
uses label vectors which we extracted using the Abstract Scenes
metadata. These label vectors encode the coordinates and rotation of
all objects and agents in images, as well as the physical position and
facial expression presented by agents. Our unsupervised pretrained
image encodings may not have learnt to extract all of the information
contained in the label vectors. Thus, the visual-labels model serves as an
upper bound in our evaluations for what semantic and syntactic
information can be learnt if given all of the visual grounding
information available.

\section{Experiment 1: Semantic bootstrapping and joint learning}\label{sec:sem-boot}

This experiment serves to answer our first research question: can access to visual-grounding and the ability to learn semantic representations in a joint learning setting facilitate learning grammars that generalize better to unseen contexts? In other words, akin to semantic bootstrapping, does having access to visual-grounding, or the ability to map visual meanings to strings, help models learn syntactic categories and productive syntactic rules over language. We compare four models, the joint-learning model, the semantics-first model, the syntax-first model, and the visual-labels model. For each model we train five versions from different random seeds. Based on Pinker's \citeyear{pinker1984language} proposal, access to semantic representations is the basis for learning meaningful syntactic categories. We thus expect models with access to semantic representations from the start to learn more meaningful grammars over language, with syntactic categories that better mirror those described in language acquisition and linguistic theories. Furthermore, we hypothesize that joint learning helps constrain the hypothesis space considered over grammars leading to less variation across model runs. We thus expect the joint-learning and the semantics-first models to learn better grammars than the syntax-first model and that the joint-learning model should show less variation across runs than the semantics-first or syntax-first models.

\subsection{Evaluations}
To determine what makes an induced grammar a \textit{better} grammar, we report two evaluations. The first follows previous work in grammar induction and compares the internal branching structure, excluding the root and leaf branches (which are deterministic), of the most likely parse for each sentence under the induced grammars to gold standard  parses. We also compare our models to fully left-branching and fully right-branching baselines. The second evaluation compares the induced pre-terminal categories of models to gold syntactic categories to determine whether the induced categories over words correspond to syntactic categories traditionally used in linguistics to describe lexical categories. To do this, we visualise these mappings in contingency tables for each model seed and measure alignment across model seeds. This second evaluation is inspired by semantic bootstrapping theory, which was intended to explain both how children learn syntactic primitives, especially formal categories over words, and what distributional information they must attend to in their input to do so \cite{pinker1984language}. With this second evaluation, we can compare the impact of visual-grounding and objective functions on the models ability to induce meaningful syntactic categories, following Pinker's proposal for a model of direct evidence of the semantic bootstrapping hypothesis.

The gold parses were extracted using the Berkeley Neural Parser (Benepar) \cite{kitaev-klein-2018-constituency, kitaev-etal-2019-multilingual}, which derives constituency parses for sentences and uses part-of-speech tags from the SpaCy library \cite{spacy2020} as lexical categories. This approach is also taken in \citeNP{shi-etal-2019-visually, zhao-titov-2020-visually}. To create a correspondence between the part-of-speech tags on gold trees and syntactic categories, we used a custom mapping provided in Appendix \ref{app:cat}. We train five different random seed runs for each model.

\subsection{Results}

\subsubsection{Evaluating syntactic structure}

\begin{table}
\caption{\label{tab:f1-test} Mean span F1 scores over test sentences by model with gold tree parses. Standard deviations in parentheses.}
\begin{center}
\begin{tabular}{lc}
\hline
\textbf{Model} & \textbf{Gold parses} \\
\hline
\textit{Right-branching}  & 0.85 (0.18) \\
\textit{Left-branching} & 0.08 (0.12) \\
\hline
Joint-Learning & \textbf{0.90 (0.16)} \\
Semantics-first & 0.75 (0.24) \\
Syntax-first & 0.42 (0.21) \\
Visual-labels &  0.87 (0.17) \\
\hline
\end{tabular}
\end{center}
\end{table}

\begin{figure}
    \centering
    \includegraphics[height=6.5cm]{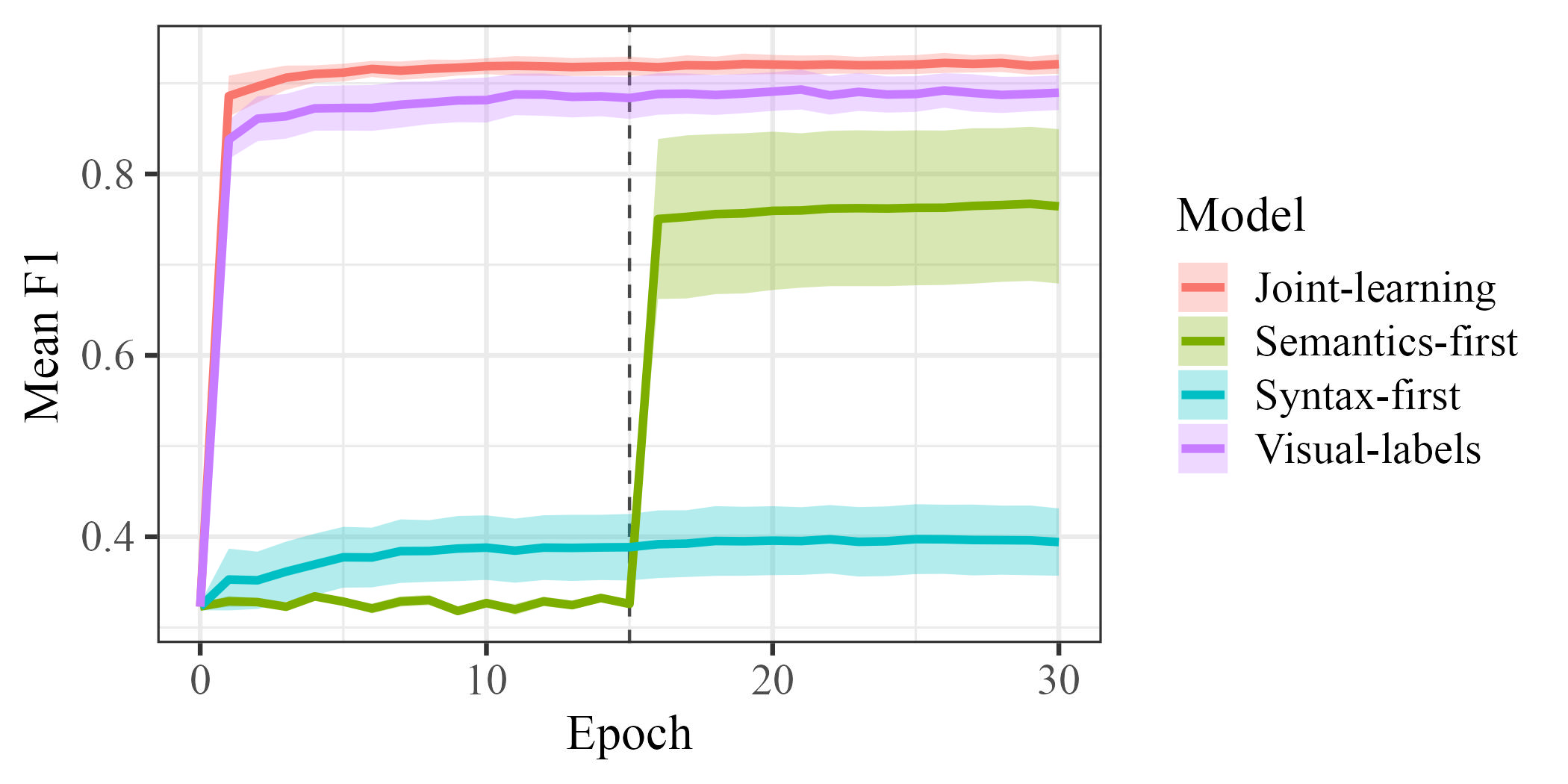}
    \caption{Mean Span F1 scores on test sentences by model during learning. Shading represents standard error across 5 runs. Dashed line represents point in time where semantics-first and syntax-first models switch to joint-learning loss function.}
    \label{fig:sem_indist}
\end{figure}

We compare the induced trees on the in-distribution test sentences of models to the gold parses, as well as fully left-branching and right-branching trees in Table \ref{tab:f1-test}. We report the mean F1 score for internal constituent spans across random seed runs and the standard deviation. The model which induces trees closest to the gold parses is the joint-learning model. Given that these are English sentences, the gold parses are quite similar to right branching trees, with on average 85\% of internal branch span correspondence.
These scores are taken after 30 epochs of learning, though in Figure \ref{fig:sem_indist}, we additionally plot the changes in mean span F1 scores between predicted trees and gold parses throughout learning for each model. The vertical dashed line in this figure marks the halfway point as which the syntax-first and semantics-first models switch to a joint-learning objective (adding in either the semantic loss or the syntactic loss respectively). Joint learning clearly leads models to successfully induce grammars over language. Whether joint learning is paired with the self-supervised image embedding or with the gold label image embeddings (visual-labels model), does not seem to matter. Interestingly, with the semantics-first model, which is simply at chance for the first half of learning since it does not yet have access to the syntax loss for learning grammar, we see that once we introduce the joint loss, it is able to learn reasonable grammars, however there is much more variation across runs. This observation supports the hypothesis that joint learning is in part most successful because it can lead to mutual constraining of the hypothesis space for both syntactic and semantic learning. As for the syntax-first model, without the availability of visual grounding initially, it induces grammars that are very different from the one used for the gold parses\footnote{Though these results may seem low, they are in line with previous findings on other corpora, where non-visually grounded C-PCFG mean F1 branching scores range between 0.36 and 0.55 \cite{kim-etal-2019-compound, zhao-titov-2020-visually}.}. Even once visual grounding is made available at the halfway point, the model can no longer recover, having limited itself to a certain hypothesis space over categories and rules that is too far from those used in deriving the gold parses.

\begin{figure}
    \centering
    \includegraphics[width=\textwidth]{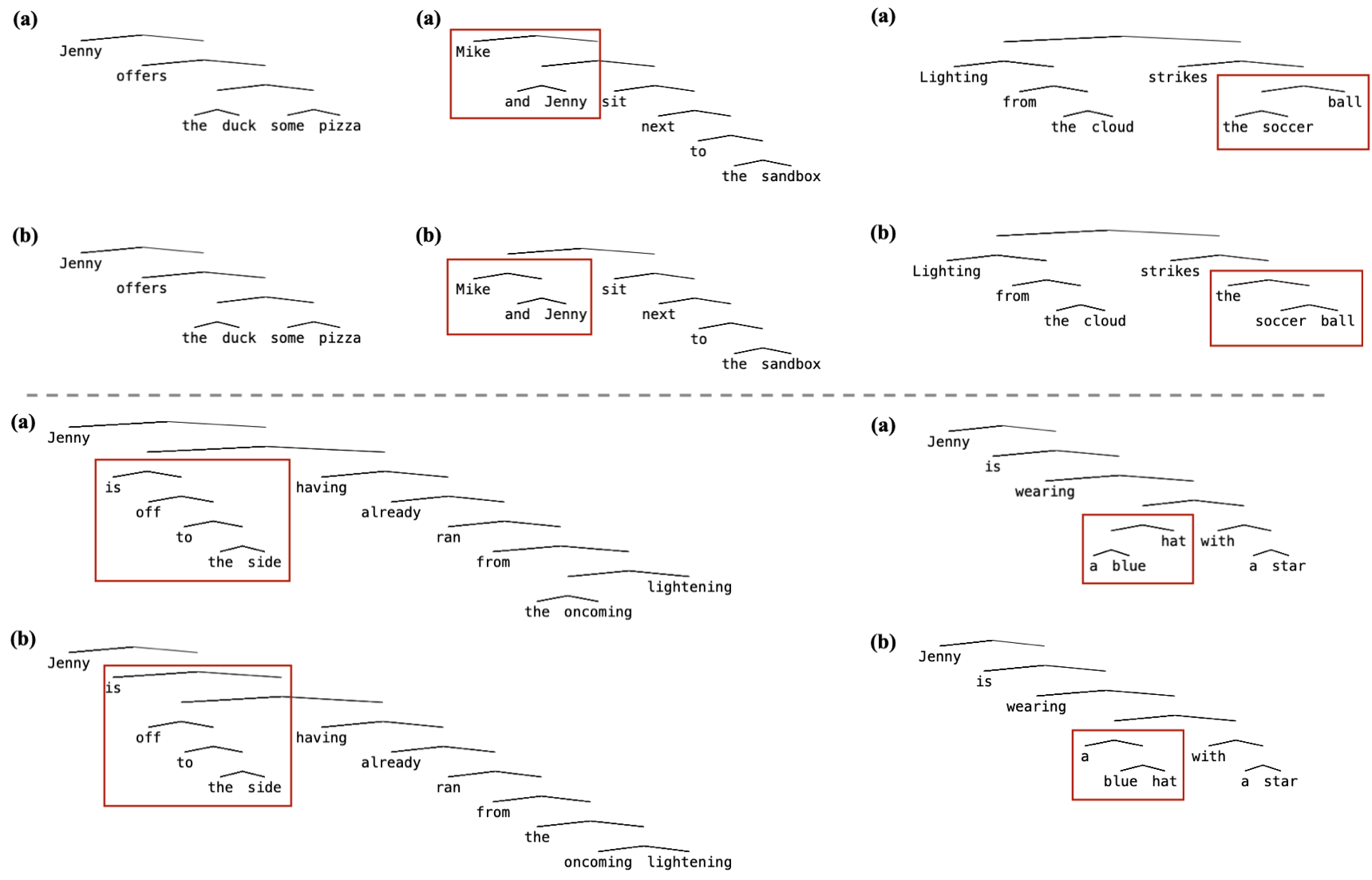}
    \caption{Examples of (a) induced trees from the joint-learning model and (b) gold parses from the Berkeley Neural Parser. Red boxes highlight discrepancies between predicted and gold trees.}
    \label{fig:pred_tree1018}
\end{figure}

In Figure \ref{fig:pred_tree1018}, we compare example induced trees from the joint-learning model to the respective gold parses to see were differences lie. One of the main difference is in noun phrases with adjectives, where we can clearly see that predicted trees have determiners subordinate to nouns, while gold parses have nouns subordinate to determiners. This difference is reminiscent of the noun phrase (NP) versus determiner phrase (DP) debates in theoretical linguistics \cite{koylu2021overview}. Another difference seems to be in the treatment of coordinate structures in subject position, where predicted trees have the first conjunct selecting for the sentence containing the second conjunct as subject, instead of having the conjunction as a whole serving as subject.

\subsubsection{Comparing induced syntactic categories}

\begin{figure}
    \centering
    \includegraphics[width=\textwidth]{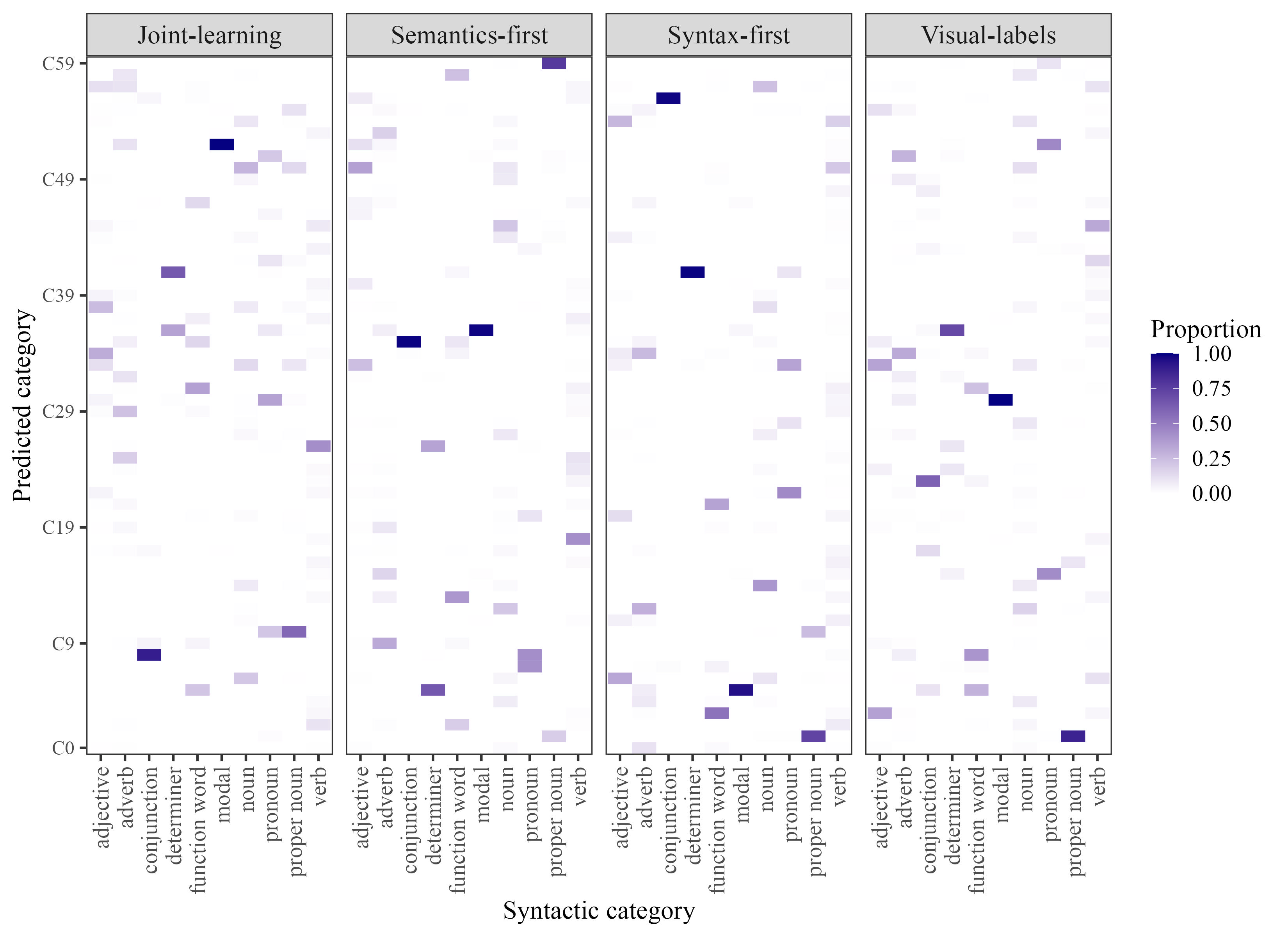}
    \caption{Proportion of predicted category to syntactic category mappings on all sentences. Models with random seed 1018 (other random seed results available in supplementary materials)}
    \label{fig:pred_cat1018}
\end{figure}

The first evaluation considered the quality of the induced grammar. In this second evaluation, we look at the pre-terminal or lexical categories having been learnt over words. We compared the induced categories for words in predicted parses, to the syntactic categories associated with words in the gold parses. Since each random seed run may learn a different mapping between categories, we first consider each run separately and visualize how well predicted categories map to gold ones using a normalized contingency table. In Figure \ref{fig:pred_cat1018}, we plot the proportion of words in each syntactic category that was mapped to a given pre-terminal category --- of which there were up to 60. The same plots for other random seeds are available in Appendix \ref{app:cat}.

With the exception of the syntax-first model, others seem to distribute the majority of words into 10 to 20 pre-terminal categories. In the case of the joint model, most syntactic categories map to distinct sets of pre-terminals, suggesting that the model was able to learn syntactically meaningful lexical categories over words. There is some overlap between nouns, proper nouns, and pronouns, though since these types of words tend to occupy similar syntactic positions, it is not too surprising. The semantics-first, syntax-first, and visual-labels models were also able to induce meaningful lexical categories. We note that the syntax-first model seems not to have any clear correspondence in its pre-terminals to verbs. Since verbs are so important to sentential structure, the lack of a clear verb category or set of categories may in part explain why this model also fails to induce grammars which resemble our gold parse grammar in the first evaluation. 

Additionally, we measure the quality of models induced lexical categories using V-measure, an entropy-based cluster evaluation measure\cite{rosenberg-hirschberg-2007-v}. Similarly to F-scores, V-measure is a weighted harmonic mean of two cluster quality metrics, homogeneity and completeness\footnote{All these measures are bounded between 0 and 1, where in general higher values are considered better.}. Intuitively, homogeneity measures how well induced lexical categories map to labelled syntactic categories (i.e. do all words in C52 map to a single syntactic category, like \textit{modal}?), while completeness measures how well labeled syntactic categories map to induced lexical categories (i.e. do all words labeled \textit{adjective} map to a single pre-terminal category, say C35?). Given that we are interested in inducing homogeneious lexical categories where words all share particular syntactic properties, this measure matters most to us. Completeness may matter less here since the level of precision of our labeled syntactic categories is arbitrary and models may in fact be inducing more precise categories --- for example, had we chosen to separate the category \textit{verb} into more subcategory labels such as transitives, intransitives, and ditransitives, it is possible that completeness would then be higher. We report mean V-measure, homogeneity, and completeness across model runs in Table \ref{tab:v-test}.

\begin{table}
\caption{\label{tab:v-test} Mean V-measure cluster evaluation results for predicted pre-terminal categories and labeled syntactic categories. V-measure is a $\beta$-weighted harmonic mean of homogeneity and completeness. Here $\beta = 0.3$ to weight homogeneity more importantly. Standard deviations are in parentheses.}
\begin{center}
\begin{tabular}{lccc}
\hline
\textbf{Model} & \textbf{V-measure} & \textbf{Homogeneity} & \textbf{Completeness} \\
\hline
Joint-Learning & 0.82 (0.01) & 0.89 (0.01) & 0.44 (0.02) \\
Semantics-first & 0.85 (0.01) & 0.90 (0.01) & 0.49 (0.03) \\
Syntax-first & 0.82 (0.01) & 0.87 (0.01) & 0.50 (0.03) \\
Visual-labels &  0.83 (0.02) & 0.90 (0.01) & 0.46 (0.04) \\
\hline
\end{tabular}
\end{center}
\end{table}

The results in Table \ref{tab:v-test} confirm that all models were able to learn meaningful lexical categories for the most part and that joint learning can lead to successful syntactic category learning. We note however that the syntax-first had the lowest homogeneity measure, while the semantics-first and visual-labels models had the highest. This observation further suggests that access to visual grounding early in learning as a proxy for semantic meaning can improve a model ability to learn syntactically relevant lexical categories, as predicted by the semantic bootstrapping theory.

\section{Experiment 2: Syntactic bootstrapping and joint learning}\label{sec:syn-boot}

In this experiment we answer our second research question: can access to linguistic structure and the ability to learn grammar in a joint learning setting facilitate learning and interpreting novel words and contexts?  Akin to syntactic bootstrapping, we would like to test if syntactic structure can help models interpret novel word meanings in context. Like in our first experiment, we compare four models: the joint-learning model, the semantics-first model, the syntax-first model, and the visual-labels model, training five versions from different random seeds. This experiment is inspired by experimental designs from studies on syntactic bootstrapping \cite{fisher2020developmental}. We hypothesise that access to syntactic structure should better a models ability to correctly map sentences to visual scenes. Joint learning should increase a model's ability to interpret novel words and contexts. Therefore, we expect joint-learning models -- both the visual-labels model and joint-learning model with self-supervised image embeddings -- to perform best, while the semantics-first and syntax-first models should see an increase in performance after introducing joint learning.

\subsection{Evaluations}

\begin{figure}
    \centering
    \includegraphics[height=5cm]{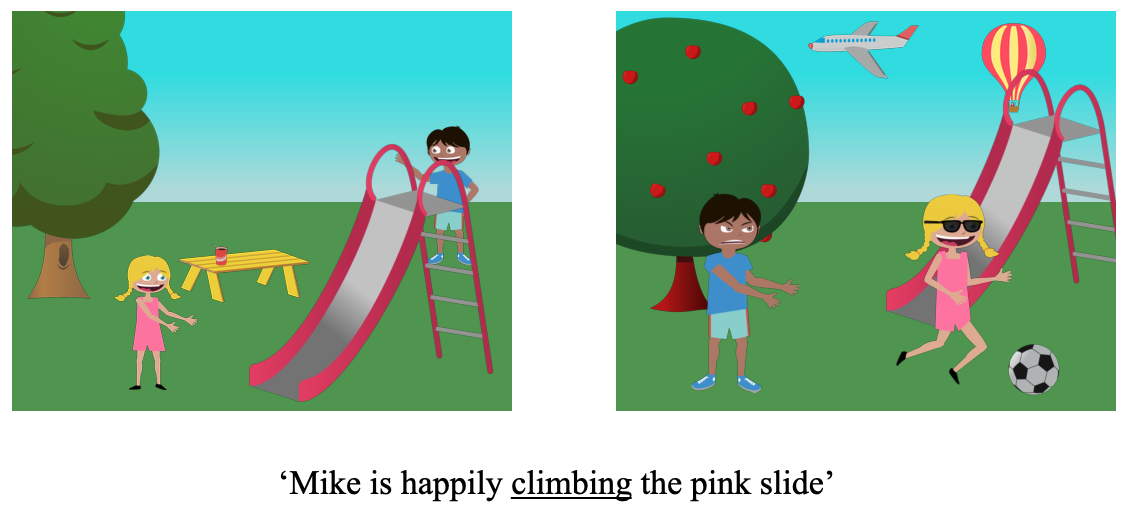}
    \caption{First evaluation: matching sentences with novel verbs to images. Example test item from transitive-inanimate condition. Models have never seen the verb to climb during learning. 
    }
    \label{fig:ex-match}
\end{figure}

\begin{figure}
    \centering
    \includegraphics[height=5.5cm]{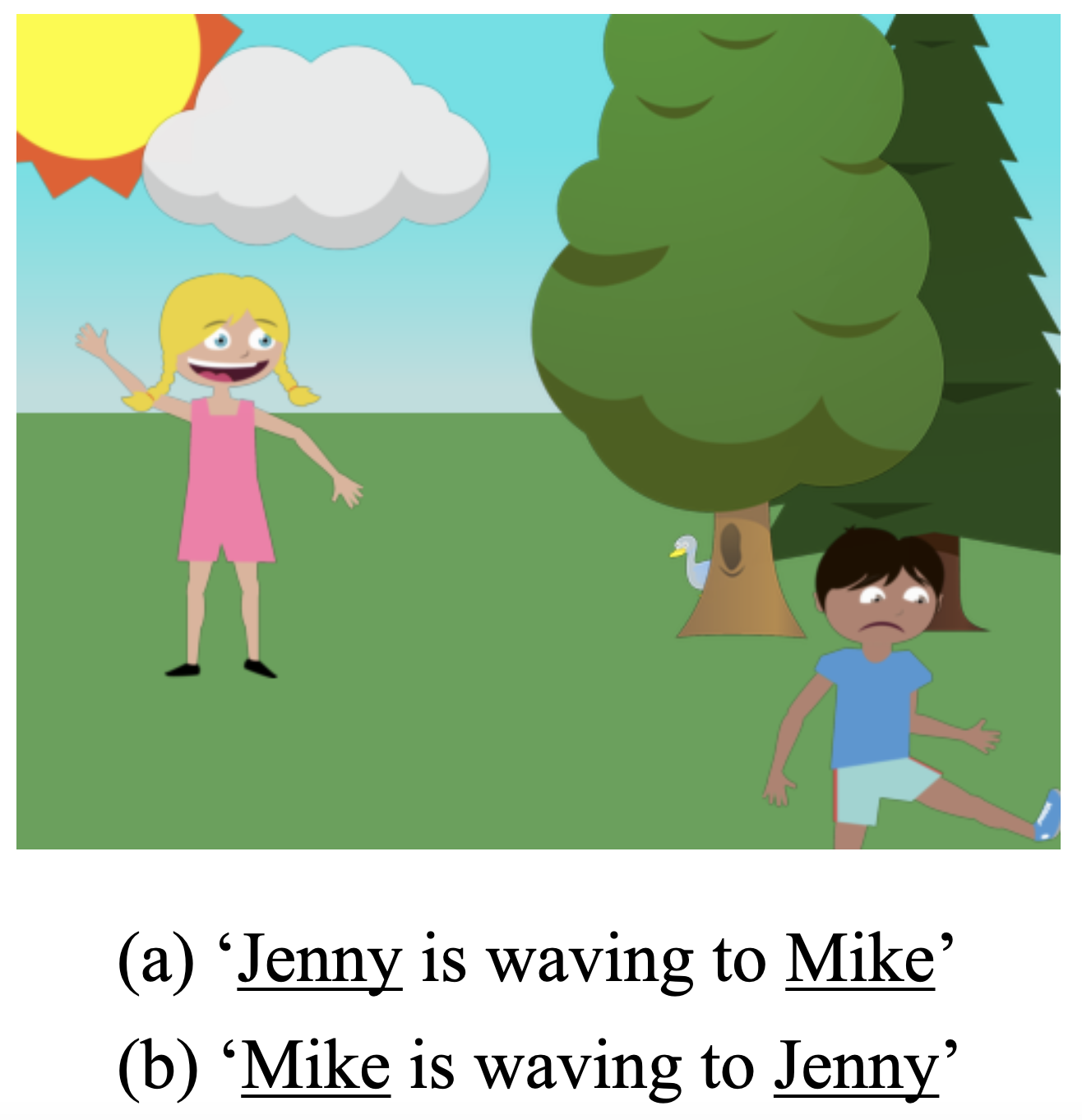}
    \caption{Second evaluation: matching semantic roles to images. Example test item with semantic role alternation. This is an in-distribution evaluation which tests whether models can distinguish semantic roles using sentence minimal pairs.}
    \label{fig:ex-match-sem}
\end{figure}

There are two evaluations for this experiment: The first involves matching a novel sentence containing a never seen verb to visual scenes (see Figure \ref{fig:ex-match}), the second involves matching a visual scene to sentences which are minimally different, only interchanging semantic roles (see Figure \ref{fig:ex-match-sem}).

The first evaluation is an out-of-distribution test done with models for which we have witheld all test sentences containing the 30 different verb stems (10 transitive-animate, 10 transitive-inanimate, 10 intransitive) from their training data. The sentences containing these held out stems served as test sentences to evaluate whether or not models could interpret these novel verbs and contexts by correctly identifying the corresponding images from a pair of images. The test examples are constructed by pairing one transitive sentence with an intransitive one, where there respective images then serve as each other's target and distractor images. This test is based on the previously mentioned nonce verb learning experimental paradigm shown in Figure \ref{fig:ex-exp}. We note that our evaluation does differ in some ways from this setup. Since our test examples are built using existing images and sentences from the Abstract scenes dataset, they are not necessarily minimal pairs, so it is possible at times to use additional sentential cues to help correctly identify the target image. For example, in Figure \ref{fig:ex-match}, models could use the presence of the adverb `happily' to help pick the correct image, since in the distractor image, Mike does not seem happy. Our evaluation setup however allows us to have many test items, 1436 to be exact. We cannot fully control for additional factors beyond the syntactic structure surrounding novel verbs contributing to models correctly matching sentences to images. For this reason, we additionally introduce a second evaluation.

Syntactic bootstrapping theory argues that children can use their understanding of argument structure and semantic roles to help them interpret and learn the meanings of novel verbs \cite{gleitman1990structural}. To determine if models have learnt to distinguish semantic roles and thus whether they could be using a similar strategy to correctly interpret novel verbs in the first evaluation, we use a second follow up evaluation taken from \cite{nikolaus-fourtassi-2021-evaluating}, illustrated in Figure  \ref{fig:ex-match-sem}. Here, models are given an image with a transitive action and two minimally different sentences, only differing in that the agent and patient roles have been reversed. Models must then correctly identify the sentence which contains the appropriate semantic role assignment based on the image. This second evaluation is an in-distribution test using carefully constructed sentence minimal pairs with known verbs that were not in the original Abstract Scenes dataset. It is more controlled and can pinpoint how much models understand about the argument structure of known verbs and their semantic roles, however it has only 50 test items.

\subsection{Results}

\subsubsection{Matching sentences with novel verbs to images}

\begin{figure}
    \centering
    \includegraphics[height=6.5cm]{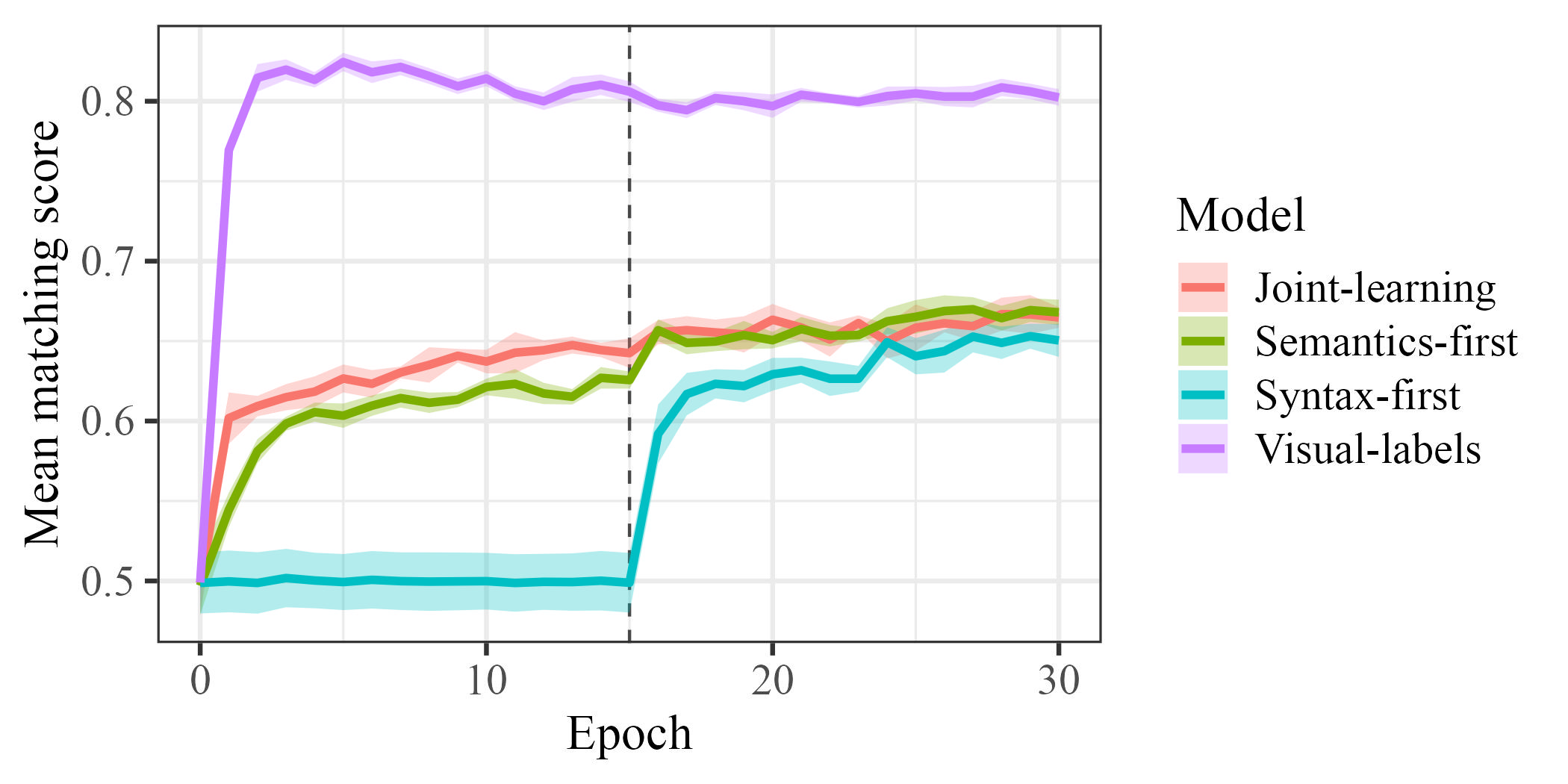}
    \caption{Matching novel verbs: Mean matching scores on sentences with out-of-distribution verbs by model during learning. Shading represents standard error across 5 runs.}
    \label{fig:syn_all_outdist}
\end{figure}

This first evaluation measures how well models can interpret sentences containing novel verbs and whether they can distinguish between transitive and intransitive actions. We report the average proportion of correctly identified target images, or the mean matching score, across 5 random seed models and 1436 test items. We plot models' matching scores throughout training in Figure \ref{fig:syn_all_outdist}. Chance performance is 0.5 since this is a balanced binary choice task. All models successfully learn to map novel verb sentences to their respective images the majority of the time. We note that the visual-labels model which has gold labels as visual encodings for image content does much better than other models. All the other models as previously described use self-supervised visual encodings which may not have learnt to encode all the necessary visual information. Still, we see that the joint-model's performance consistently increases while both the syntax-first (at chance initially since it has no access to visual encodings) and the semantics-first models see a jump in performance after the introduction of the joint-learning loss at the half way point\footnote{Astute readers will have noticed that though the syntax-first model saw no improvement after the introduction of joint learning in experiment 1 on semantic bootstrapping in section \ref{sec:sem-boot}, here the syntax-first model does catch up after the introduction of joint learning. This observation suggests that is not so much the particular verb phrase structure we learn for transitive and intransitive verbs that matters for syntactic bootstrapping (e.g. ``[Agent [verb [Patient]]]" and ``[Agent [verb]]"), but simply that they be distinguishable, for example ``[[[Agent] verb] [Patient]]" versus ``[[Agent] verb]", may not correspond to our gold parses, but they still make a distinction between verb type structures and may have relevant constituents to map to semantic representations.}. Performing a paired t-test for the semantics-first model right before and after having introduced joint learning confirms that this jump is significant (\textit{t}([4]) = [5.04], \textit{p} = [> 0.01]).

\begin{figure}
    \centering
    \includegraphics[height=6.5cm]{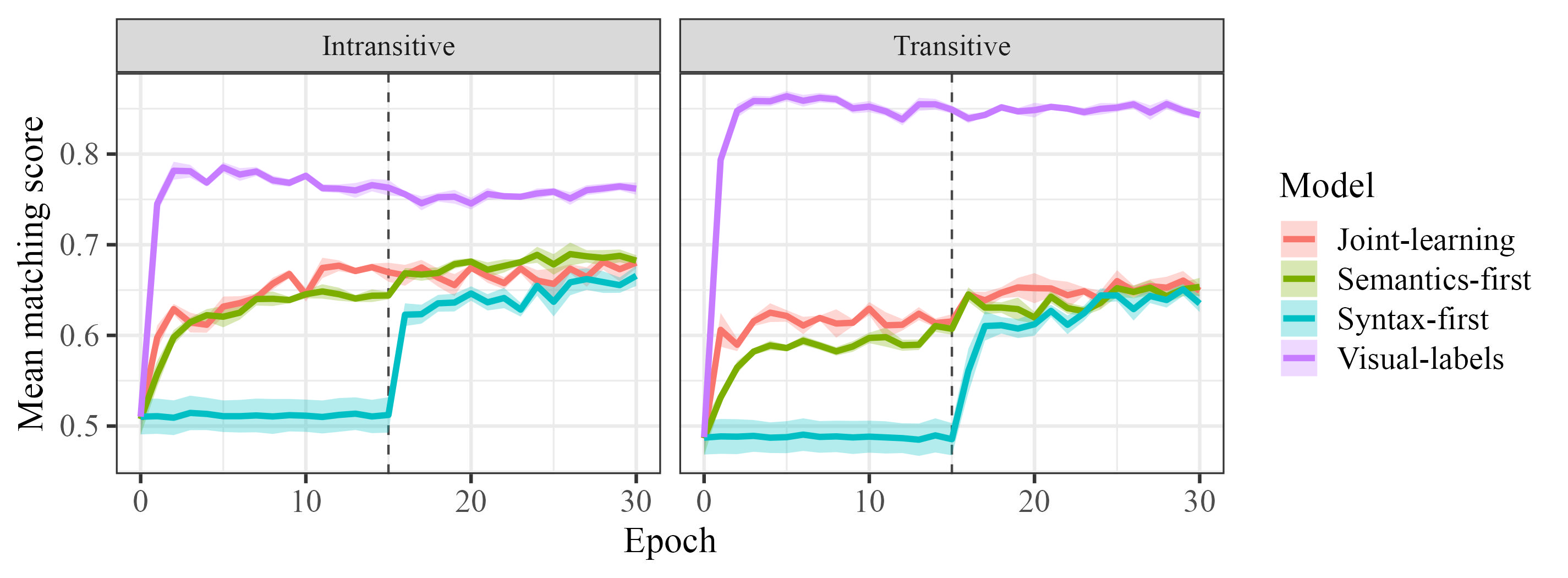}
    \caption{Matching novel verbs: Mean matching scores on out-of-distribution sentences by verb type and by model during learning. Shading represents standard error across 5 runs.}
    \label{fig:syn_vtype_outdist}
\end{figure}

We additionally plot model performance with respect to target novel verbs being either transitive or intransitive conditions in Figure \ref{fig:syn_vtype_outdist}, each with 718 test items. We see no meaningful difference between conditions for the joint-learning model and syntax-first model. The jump in performance observed in the previous figure for semantic-first models seems to be in the transitive condition more prominently. Transitive sentences contain more structure and require more mastery of semantic roles to properly interpret them; access to syntactic information via joint learning may be especially important in this condition, showing evidence of model behaviour that is consistent with the syntactic bootstrapping hypothesis. Interestingly, the visual-labels model struggles more in the intransitive condition than the transitive condition. Since this difference is not present in the other models, this observation likely indicates that the visual-labels model is relying more heavily on the additional visual information in some contexts over syntactic information. Since transitive sentences and their corresponding images are likely to contain more distinct referents, especially in the transitive-inanimate object condition, the difference in performance between these two conditions could be explained by the model relying on distinctive visual cues (eg. the presence of a soccer ball in one image versus another) in some contexts. The same does not seem to be true of other models.

\begin{figure}
    \centering
    \includegraphics[height=6.5cm]{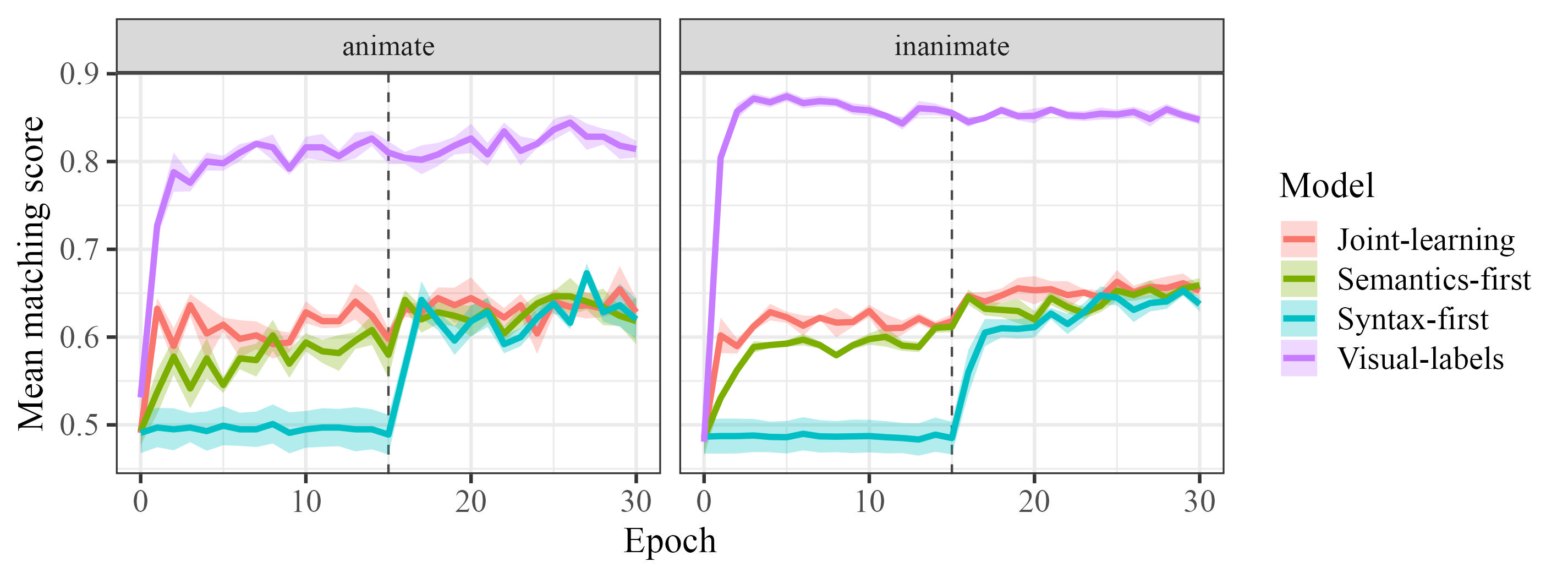}
    \caption{Matching novel verbs: Mean matching scores on out-of-distribution transitive sentences by object type and by model during learning. Shading represents standard error across 5 runs.}
    \label{fig:syn_tr_obj_outdist}
\end{figure}

Within the transitive condition, we compare performance between trials that had an animate (person or animal) or inanimate object in Figure \ref{fig:syn_tr_obj_outdist}. We find more evidence in favor of our hypothesis about the visual-labels model, as it performs a little better in the inanimate condition which contains more distinctive referents. While other models once again show little difference, learning to correctly respond the majority of the time regardless of object type. The curves for transitive animate are slightly noisier because there are fewer examples of these in the test set, 99 over 618 for the inanimate object examples\footnote{Transitive verbs taking inanimate objects are generally much more frequent in the corpus.}.

\subsubsection{Matching semantic roles to images}

\begin{figure}
    \centering
    \includegraphics[height=6.5cm]{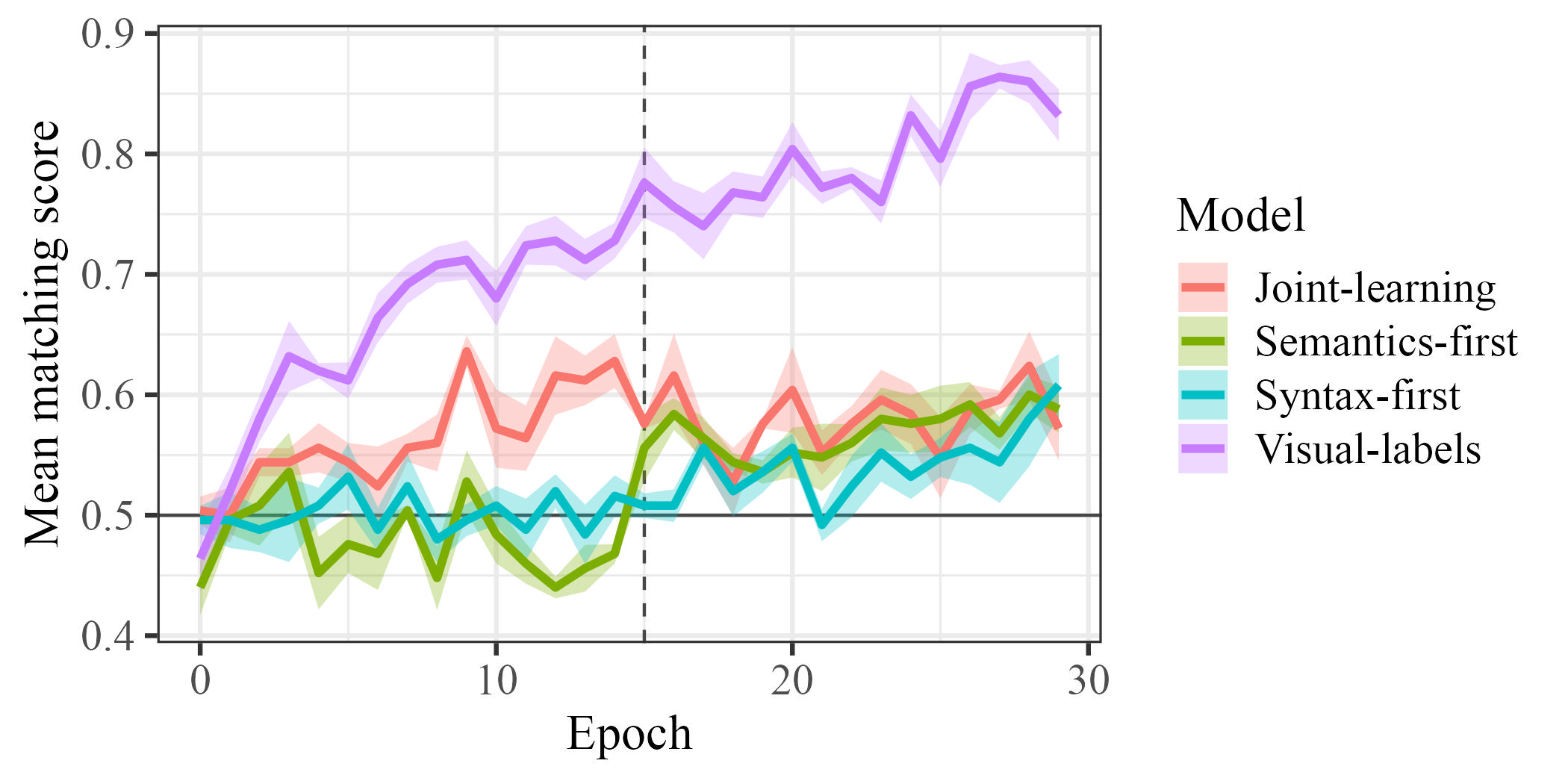}
    \caption{Matching semantic roles: Mean matching scores on semantic role test sentences by model during learning. Shading represents standard error across 5 runs.}
    \label{fig:sem_roles_indist}
\end{figure}

To determine if models are sensitive to semantic roles, we measure their ability to distinguish between minimally different sentences containing reversed semantic roles given an image. We report the average proportion of correct matches across random seed runs and 50 test items, or models' mean matching score. Chance performance is once again 0.5. Figure \ref{fig:sem_roles_indist} shows models' performance throughout learning. Neither the syntax-first and semantics first models do much better than chance at first. This pattern is expected for the syntax-first model, but for the semantics-first model, it highlights the importance of syntactic structure for learning semantic roles. Once joint learning is introduced at the half-way point their performance starts to increase. The joint-learning model and visual-labels model see a consistent rise in performance, which plateaus around the half way point for the joint-learning model, likely due to the limitations of its visual encodings.

Given the limited number of test examples, the performance curves are quite noisy. It is clear that this task is difficult and that all models but the visual-labels model are limited by the accuracy of their visual encoders. Still, joint learning allows models to successfully learn to identify semantic roles in a many cases, and when not limited by visual perception, in most cases.

\section{Discussion}

We set out to show via a computational cognitive model that both semantic bootstrapping and syntactic bootstrapping effects arise as a result of the interplay between syntactic and semantic knowledge acquisition during joint learning. Semantic bootstrapping and syntactic bootstrapping are not necessarily meant to be mutually exclusive and in fact, as our experiments suggest, the strongest effects of syntactic and semantic bootstrapping arise when we view language learning as a joint inference problem, where both semantics and syntax are learnt simultaneously.

Akin to semantic bootstrapping, existing work on neural visually/semantically grounded grammar induction \cite{shi-etal-2019-visually, zhao-titov-2020-visually, jin-schuler-2020-grounded, wan2022unsupervised, li2024reevaluating} have found that access to images or large language model semantic embeddings can lead to moderate improvements in grammar induction; in this work, we found that our visually-grounded joint-learning model in fact saw large improvements, learning much better grammars than a syntax-first or syntax-only model. Our large improvements over moderate ones seen in previous work may be due to the following differences. To start, other studies used image-caption datasets (e.g. MS-COCO ; \citeNP{lin2014microsoft, chen2015microsoft}) where captions are not complete sentences with main verbs for the most part, which may be important when considering grammar acquisition as a whole. Furthermore, our model not only learnt the grammar from scratch, but also its visual and semantic representations, such that, akin to syntactic bootstrapping, learning semantic representations could take into account syntactic relations between referents. A final difference and possible limitation to our study is that we used synthetic images while previous studies used real world images. It is possible that synthetic images made the task of identifying visual feature easier. Children's input is undoubtedly noisier and richer than the data our model was exposed to, but image-caption datasets are no closer to their learning experience either. Given our primary goal: to demonstrate how the interplay between syntactic and semantic acquisition can follow from joint learning, using simulated children's book data presented a sufficient environment to examine these dynamics.

Joint learning works because it helps mutually constrain related hypothesis spaces, here grammar and semantic representations. These types of constraints are likely necessary for human learners who---unlike large language models---are limited in terms of memory and processing capacity as well as amount of input evidence. Even with these limitations though, we learn language and better yet, we learn representations which allow to generalize and use language in completely novel contexts. The reason for our learning efficiency and generalization abilities may lie in our effective learning strategies, which we argue are built on joint learning.

Empirical evidence suggests that semantic and syntactic processing during language comprehension or production are not separable into distinct areas of the brain, but instead represent distributed processes which overlap across a wide region referred to as the language network \cite{fedorenko2020lack, hu2022precision, shainetal23:nlength,fedorenko2024language, shain2024distributed}. Furthermore, children's lexicon and their syntactic production abilities grow side by side during language development \cite{bates1994developmental, brinchmann2019there, frankToAppearWordbank}. These results all support our proposal: that language learning is joint learning across many levels of linguistic representation. The acquisition of morphemes, words, syntax, semantics, pragmatics have for the most part been considered in isolation. However, if language learning is indeed a joint inference problem across many levels of linguistic structure, then future research in the field should try to understand \textit{how} learning biases or constraints within these different levels arise as a function of joint learning. For example, how does the acquisition of semantic knowledge affect the acquisition of syntax? or how does learning morpheme boundaries interplay with the acquisition of semantic knowledge? Understanding how these constraints arise and interact, we suggest should be the next key direction in language learning debates.

Computational modeling is not new to the fields of language development and cognitive science. However, the ability to design models like the one in this paper as well as the access we now have to multimodal data---sound, image, video, text--- may allow us to revisit with new perspective many of the research questions we still have about language learning. We have proposed that a promising way to do so is to think of language learning as a holistic problem involving joint inference over many different levels of abstract linguistic representation. We have shown how joint learning does not necessarily make learning \textit{harder}, but can make it \textit{easier} by mutually constraining the hypotheses being considered by a learner, helping them acquire the complex systems that are human languages. We hope that this work may convince other researchers in both cognitive science and AI that an important new direction for language modeling and learning research lies in considering the dynamics of joint inference over many input sources and modalities as well as levels of representation.

\newpage
\bibliographystyle{apacite}
\bibliography{main.bbl}

\appendix
\section{Verb stem lists for data split} \label{app:data}
These are the stems of verbs which appeared at least 5 times in the corpus. We held out for testing all instances of the following verb stems:

\paragraph{Held out intransitive verb stems} \textit{taking animate objects} -- push, rescu, teas, argu, hug, warn, feed, meet, fight, invit -- \textit{taking inanimate objects} -- drop, open, ride, pour, brought, prepar, toss, use, climb, rais (hand).
\paragraph{Held out transitive verbs} walk, hide, smile, cheer, laugh, slid, cri, danc, fell, crawl

The training data included verbs with the following stems:

\paragraph{Training verb stems} is, wear, are, sit,  hold, has, stand, play, want, fli, slide, hot, kick, run, sad, scare, swing, mad, rain, wave, picnic, grill, bat, see, catch, watch, go, throw, jump, afraid, look, was, tri, surpris, eat, set, will, threw, drink, upset, have, get, excit, like, come, hand, shine, worri, doe, be, hit, chase, cook, made, camp, had, talk, fall, wait, give, start, think, sat, float, put, got, hurt, help, took, wore, saw, growl, flew, were, call, love, lost, went, shock, carri, offer, make, take, roar, pet, toy, found, stole, came, let, land, enjoy, can, tell, yell, pretend, reach, slither, strike, startl, burn, say, would, fetch, shin, stuck, know, ran, caught, did, goe, move, stare, find, follow, could, share, do, bring, might, rest, show, leav, round, notic, built, feel, waiv, ruin, perch, grow, pick, frighten, stop, miss, bake, struck, warm, seem, scream, leg, cover, lay, thunder, snif, keep, stay, should, color, ask, fallen, cross, bite, face, held, said, done, blast, roll, pass, ate, bounc, taken, ripe, hate, gone, thrown, closer, gave, place, stood, seen, tie, care, wish, point, hope, begin, pitch, forgot, holdng, waddl, snuck, annoy, tire, steal, dig, barbecu, dri, wonder, shout, been, decid, soar, skip, frown, understand, glad, dress, approach, sneak, shade, lose, copi, alarm, build, finish, astonish, standng, block, touch, knock, amus, plan, confus, better, may, hover, stripe, jog, told, bore, need, head, listen, thrill, join, began, attack, trade, smell, grab, trip, frustrat, stit, storm, stolen, hidden, sleep, sail, snake, hear, kneel, launch, march, juggl, serv, protect, site, tast.

\section{Model implementation} \label{app:model}

All models were trained on A100 Multi-Instance GPU partitions, using at most 32Gb of GPU memory. Each model took about 18 hours to train and evaluate at each epoch, for 30 epochs.

\subsection{$f_s$ and $f_t$ syntactic category MLPs}
$f_s$ and $f_t$ are both MLPs with a linear input layer, two layers with \texttt{ReLU} non-linear activations, and finally an output linear layer. Their only difference is that the final linear layer output is either over non-terminal symbols or the vocabulary respectively.

\subsection{Variational posterior model}
In practice the variational posterior is given by a diagonal Gaussian where the mean and log-variance vectors are given by another biLSTM with a maxpooled linear output layer over the hidden states, for $\mathbf{z}$ over each batch of sentences.

\section{Syntactic category mappings}\label{app:cat}

POS tags from SpaCy were mapped to the following syntactic categories using this correspondences in Table \ref{tab:pos-syn}.
\begin{table}
\caption{\label{tab:pos-syn} Correspondence between syntactic categories and part-of-speech tags}
\begin{center}
\begin{tabular}{ll}
\hline
\textbf{Syntactic category} & \textbf{Part-of-speech tags} \\
\hline
verb & VB, VBD, VBG, VBN, VBP, VBZ \\
adjective & JJ \\
adverb & RB, RBR, RP, RBS \\
noun & NN, NNS \\
proper noun & NNP, NNPS \\
pronoun & PRP, PRP\$ \\
determiner & DT \\
conjunction & CC \\
modal & MD \\
other function word & IN \\
\hline
\end{tabular}
\end{center}
\end{table}

\begin{figure}
    \centering
    \includegraphics[width=\textwidth]{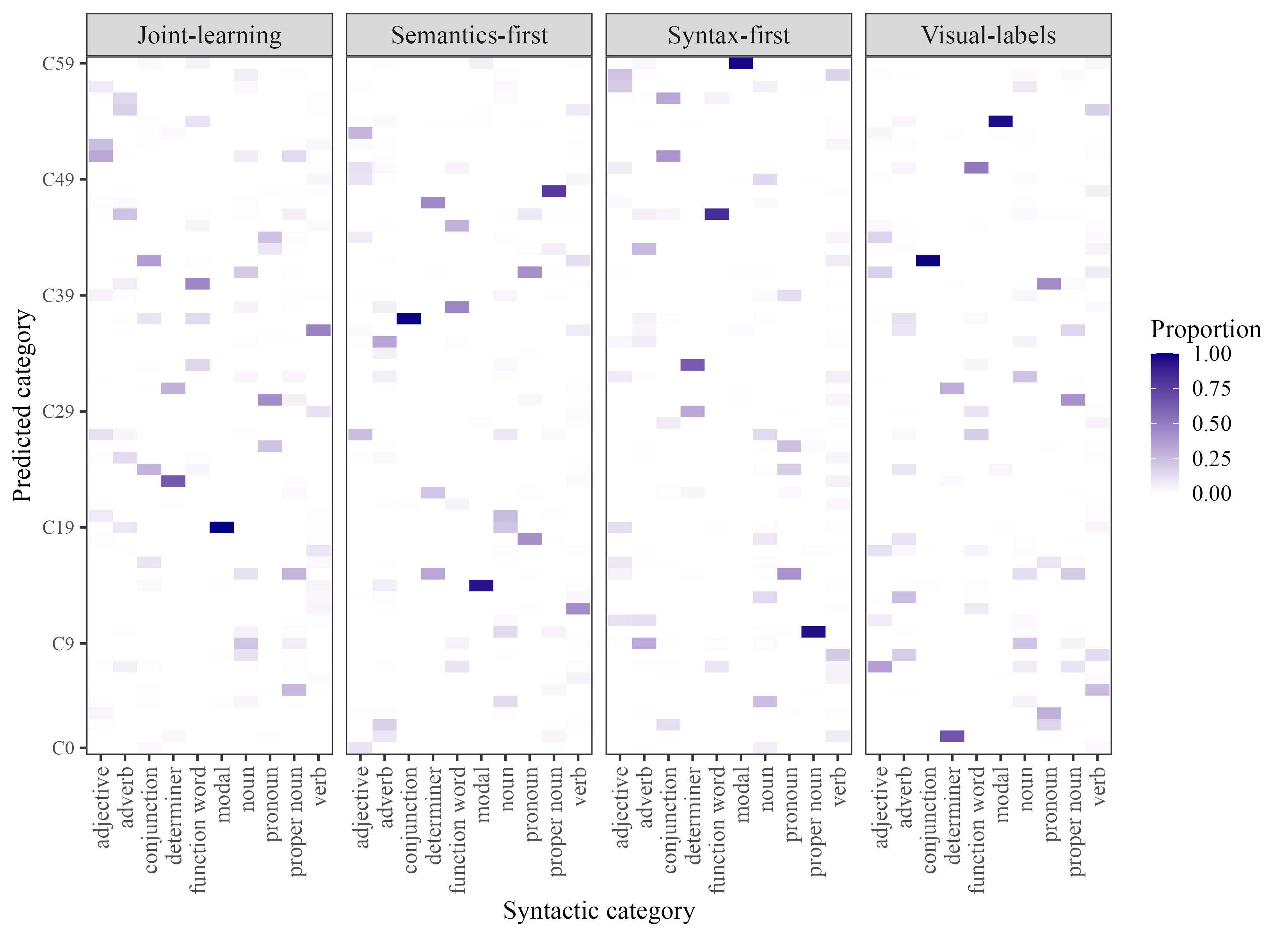}
    \caption{Proportion of predicted category to syntactic category mappings on all sentences for models with random seed 214}
    \label{fig:pred_cat214}
\end{figure}

\begin{figure}
    \centering
    \includegraphics[width=\textwidth]{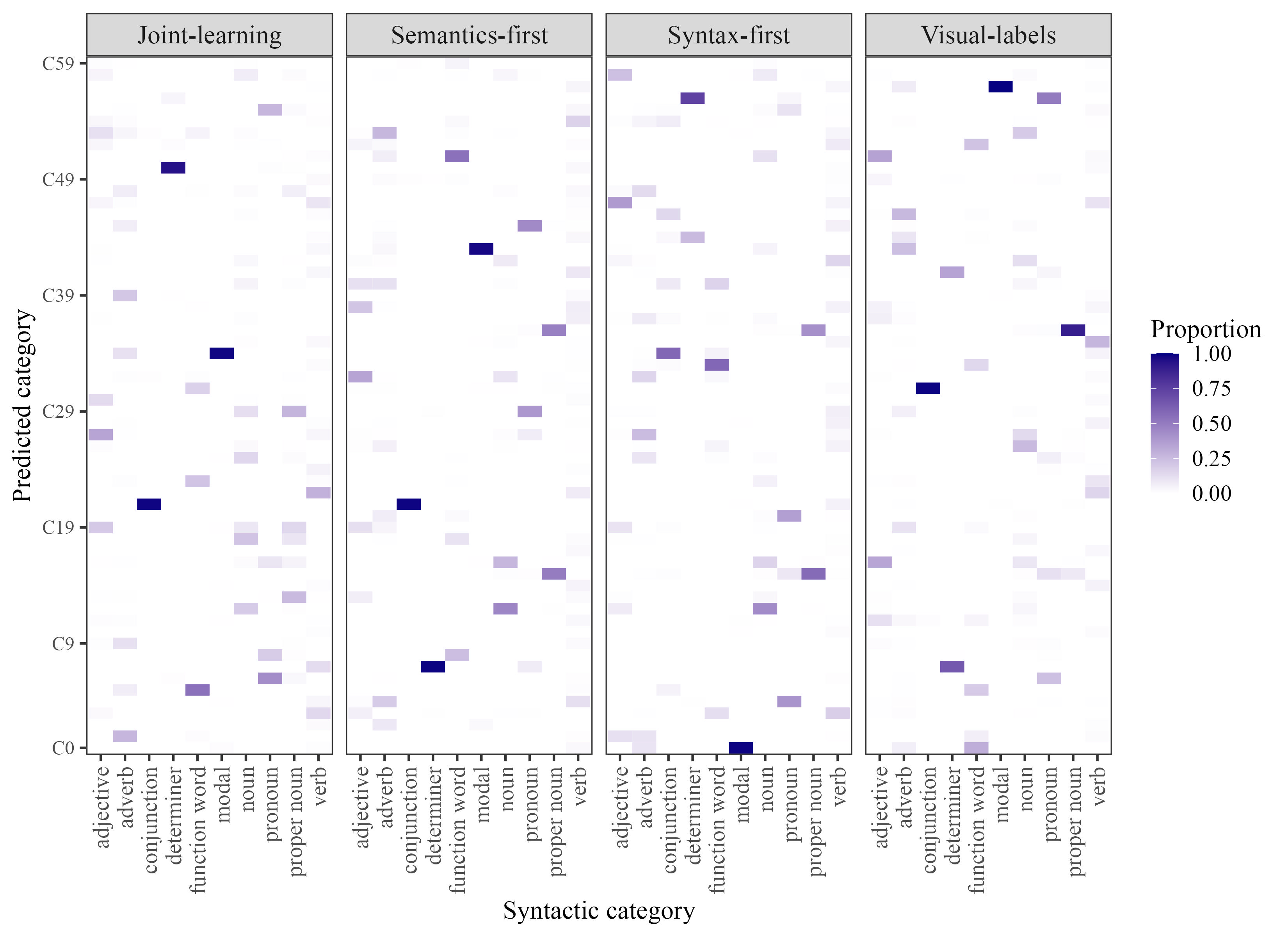}
    \caption{Proportion of predicted category to syntactic category mappings on all sentences for models with random seed 527}
    \label{fig:pred_cat527}
\end{figure}

\begin{figure}
    \centering
    \includegraphics[width=\textwidth]{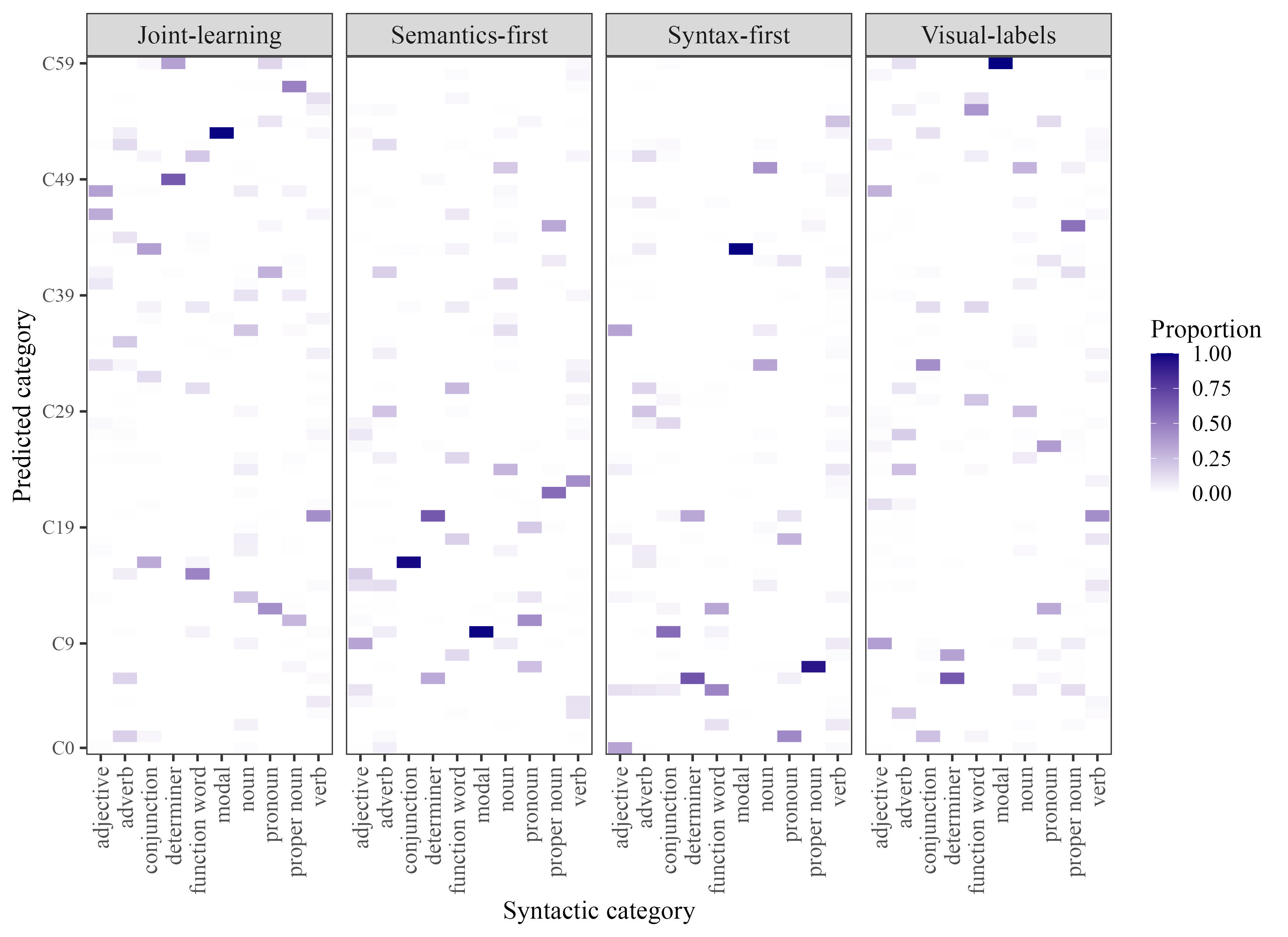}
    \caption{Proportion of predicted category to syntactic category mappings on all sentences for models with random seed 627}
    \label{fig:pred_cat627}
\end{figure}

\begin{figure}
    \centering
    \includegraphics[width=\textwidth]{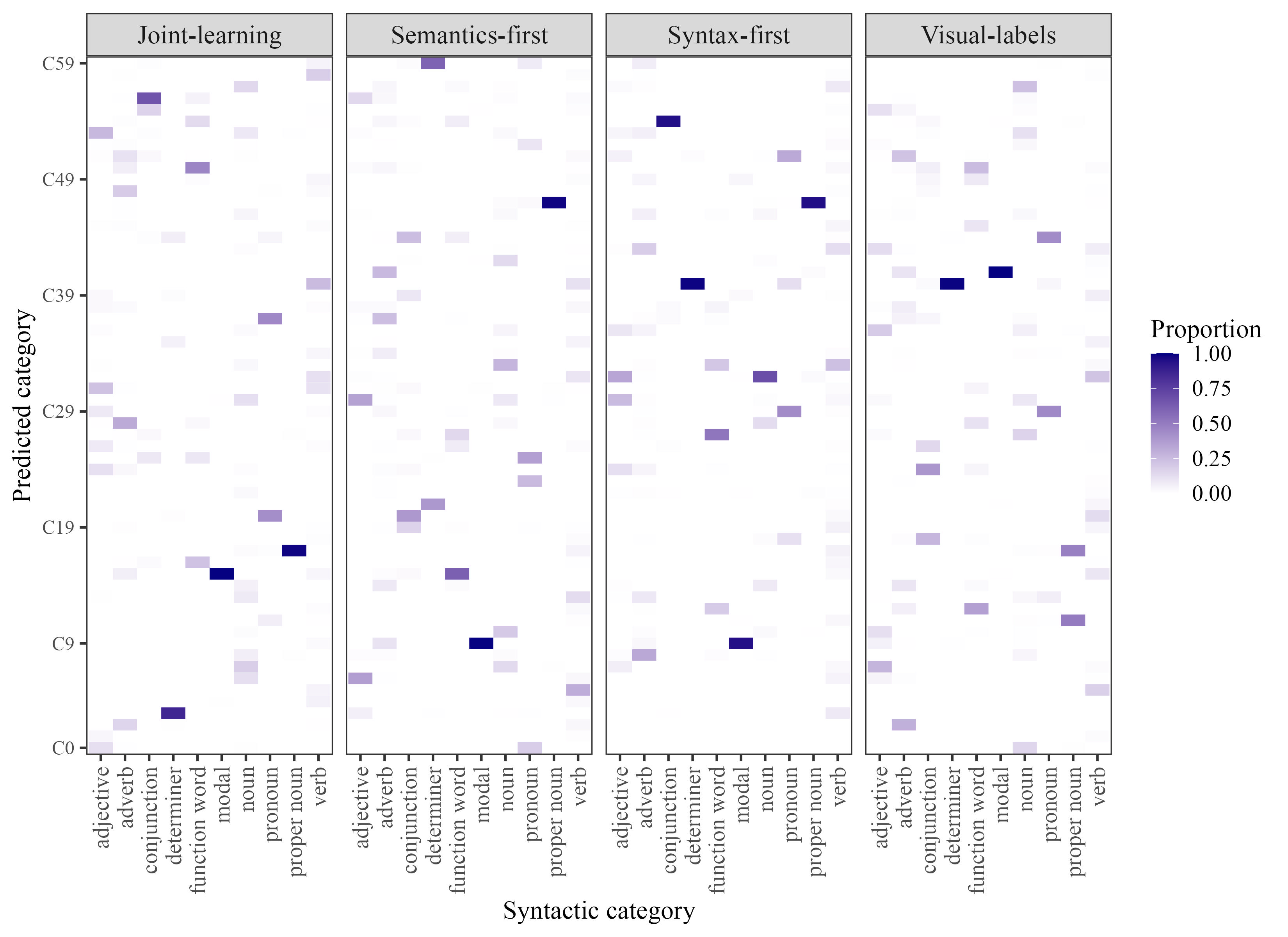}
    \caption{Proportion of predicted category to syntactic category mappings on all sentences for models with random seed 91}
    \label{fig:pred_cat91}
\end{figure}

Figures \ref{fig:pred_cat214} through \ref{fig:pred_cat91} are the predicted category to syntactic category mappings for all other model runs.

\end{document}